\documentclass[lettersize,journal]{IEEEtran}
\usepackage{algorithmic}
\usepackage[utf8]{inputenc}
\usepackage[left=15mm, right=15mm, top=20mm, bottom=20mm]{geometry} % set the margin of the manuscript

\usepackage[dvipsnames]{xcolor}
\usepackage{tikz}
\usepackage{footmisc}
\usepackage{graphicx}
\usetikzlibrary{positioning}
\usepackage{amssymb,amsmath,amsfonts,mathtools}
\usepackage{xfrac}
\usepackage{bbm}
\DeclareMathOperator*{\argmin}{arg\,min} %argmin
\usepackage{cite} %citations
\usepackage{multicol} %images spanning twocolumns
\usepackage{setspace} %double spacing
\usepackage{multirow}
\usepackage[normalem]{ulem}
\usepackage{caption}
\usepackage{subcaption}
\usepackage{url}
\usepackage{lineno}
\usepackage{tablefootnote}
\usepackage{arydshln}
\graphicspath{ {Figures/} }

\newcommand{\hs}{HS}
\newcommand{\dl}{DL}
\newcommand{\sgp}{SP}
\newcommand{\arch}{HUA}
\newcommand{\hsi}{HSI}
\newcommand{\hu}{HU}
\newcommand{\lmm}{LMM}
\newcommand{\sups}[1]{\rlap{\textsuperscript{\textbf{#1}}}}
\newcommand{\scalar}[1]{$#1$}
\newcommand{\mat}[1]{$\mathbf{#1}$}
\newcommand{\elhalf}[0]{L\textsubscript{$\sfrac{1}{2}$} NMF}
\newcommand{\udas}[0]{uDAS}
\newcommand{\cnnaeu}[0]{CNNAEU}

\newcommand{\daen}[0]{DAEN}

\newcommand{\daeu}[0]{DAEU}
\newcommand{\egu}[0]{EGUnet}

\newcommand{\seq}{GAUSS}
\newcommand{\comb}{GAUSS\textsubscript{\textit{blind}}}
\newcommand{\sig}{GAUSS\textsubscript{\textit{prime}}}
\newcommand{\garc}{GAUSS\textsubscript{\textit{ architecture}}}

\newcommand{\sref}[1]{Section \ref{#1}}
\newcommand{\fref}[1]{Fig. \ref{#1}}
\newcommand{\tref}[1]{Table \ref{#1}}
\newcommand{\eref}[1]{\ref{#1}}

\newcommand{\mapfont}[1]{\fontsize{#1}{#1}\selectfont}

\begin{document}
\bstctlcite{IEEEexample:BSTcontrol} %avoids displaying "____" for repeated authors
\title{GAUSS: Guided Encoder - Decoder Architecture for Hyperspectral Unmixing with Spatial Smoothness}
\author{D.Y.L.~Ranasinghe,
        H.M.H.K.~Weerasooriya,
        G.M.R.I.~Godaliyadda,~\IEEEmembership{Senior~Member,~IEEE,}
        M.P.B.~Ekanayake,~\IEEEmembership{Senior~Member,~IEEE,}
        H.M.V.R.~Herath,~\IEEEmembership{Senior~Member,~IEEE},
        D.~Jayasundara,
        L.~Ramanayake,
        N. Senarath
        ~and~D. Wickramasinghe

\thanks{D.Y.L. Ranasinghe, H.M.H.K. Weerasooriya, G.M.R.I. Godaliyadda, H.M.V.R. Herath, M.P.B. Ekanayake, D. Jayasundara, L. Ramanayake, N. Senarath and D. Wickramasinghe are with the Department of Electrical and Electronic Engineering, University of Peradeniya, KY 20400, Sri Lanka (email: e14273@eng.pdn.ac.lk, kavingaweerasooriya@eng.pdn.ac.lk, roshangodd@ee.pdn.ac.lk, vijitha@eng.pdn.ac.lk, mpb.ekanayake@ee.pdn.ac.lk, e15157@eng.pdn.ac.lk, lakshitharamanayake@eng.pdn.ac.lk, neranjansenarath95@eng.pdn.ac.lk and dulanthawickramasinghe001@eng.pdn.ac.lk).}%

}
\maketitle

% \linenumbers

\begin{abstract}
In recent hyperspectral unmixing (\hu) literature, the application of deep learning (\dl) has become more prominent, especially with the autoencoder (AE) architecture. We propose a split architecture and use a pseudo-ground truth for abundances to guide the `unmixing network' (UN) optimization. Preceding the UN, an `approximation network' (AN) is proposed, which will improve the association between the centre pixel and its neighbourhood. Hence, it will accentuate spatial correlation in the abundances as its output is the input to the UN and the reference for the `mixing network' (MN). In the Guided Encoder-Decoder Architecture for Hyperspectral Unmixing with Spatial Smoothness (\seq), we use one-hot encoded abundances generated through the k-means algorithm as the pseudo-ground truth to guide the UN in the initial state. Furthermore, we relieve the single-layer constraint on the MN by introducing the UN-generated abundances, in contrast to the standard AE used for \hu. In the second stage of training, we introduce two variants on the pre-trained network using the \seq\ method. In \comb, we have concatenated the UN and the MN to back-propagate the reconstruction error gradients to the encoder. Then, in the \sig, reliable abundance results of a signal processing (\sgp) method were used as the pseudo-ground truth with the \garc\. According to quantitative and graphical results for four experimental datasets, the two \seq\ variants either transcended or equated the performance of existing \hu\ algorithms from both \dl\ and \sgp\ domains.
\end{abstract}


\section{Introduction}
Hyperspectral (\hs) imaging is extensively used in earth observation satellites for various applications \cite{khan2018modern}: lithological mapping \cite{liu2018mineral, lorenz2018long, kruse2003comparison, ni2020mineral, chen2019sulfate}, environmental monitoring \cite{stuart2019hyperspectral, zeng2019multi, moroni2013hyperspectral}, and agricultural activities \cite{lu2020recent, zhu2020application, teke2013short}, since it captures more spectral information than its counterparts: multispectral and RGB imaging. However, \hs\ images (\hsi s) are of poor spatial resolution and are corrupted by noise and interference from atmospheric alterations and instrumentation imperfections\cite{rasti2018noise}. Therefore, identifying the composition of a topography requires information recovery algorithms, known as unmixing algorithms.

\hs\ unmixing (\hu) is a source separation technique that strives to extract the constituent source signal spectra and estimate their fractional presence in each pixel, referred to as endmembers and abundances. Over the evolution of \hu\ techniques, numerous unmixing algorithms have been proposed for both linear and non-linear mixing models \cite{heylen2014review, dobigeon2013nonlinear, halimi2011nonlinear, gu2021nonlinear} under signal processing (\sgp) and deep learning (\dl) domains. \sgp\ unmixing benefits from various algorithmic approaches \cite{bioucas2012hyperspectral} such as geometric \cite{nascimento2005vertex,boardman1993automating, winter1999n, miao2007endmember, zare2007sparsity, gader2012sparsity, zhuang2019regularization, li2016robust, bioucas2009variable}, statistical \cite{parra1999unmixing, dobigeon2009joint, nascimento2007hyperspectral}, and sparse regression-based \cite{bioucas2010alternating,iordache2012total}. In addition, non-negative matrix factorization (NMF) \cite{seung2000algorithms} has been profoundly used in \hu\ for its concurrent extraction of endmembers and abundances with matrix decomposition and its adherence to fundamental constraints of the \hu\ problem imposed by practical realities such having non-negative abundances. Variants of the NMF algorithm are present in the literature, which promote sparsity of the abundances \cite{qian2010l1, he2017total}, independence of the endmembers \cite{ekanayake2021constrained}, spatial-spectral information and piece-wise smoothness \cite{sun2018constraint, yang2015geometric, rathnayake2020graph, zhou2020subspace}. 

Application of \dl, though nascent, is now prominent in \hu\ due to the growing capacity in computational power, the ability to incorporate any contextual feature readily and the potentiality to extract information from unstructured data \cite{signoroni2019deep}. \dl\ unmixing is frequently performed through semi-supervised or unsupervised fashion using convolutional networks \cite{qi2020deep, zhang2018hyperspectral, palsson2020convolutional, ranasinghe2020convolutional, wang2019nonlinear, zhao2021hyperspectral, zhao20223d, 9709848} , dense networks \cite{ozkan2018endnet, ozkan2018endnet, qu2018udas, su2019daen, han2020deep, hong2021endmember}, and recurrent networks \cite{zhao2021lstm} with the assistance of other functional layers such as pooling, regularization, and shaping. Nonetheless, most \dl\ unmixing algorithms are structured as autoencoders (AEs) due to their resemblance to matrix decomposition applied in blind source separation techniques and their unsupervised learning capabilities.

However, AEs have limitations on the architecture when used for blind unmixing, even though AEs are widely employed in \dl\ unmixing. In particular,
\begin{itemize}
    \item The encoder optimization is conditioned on the decoder optimization with the end-to-end structure.
    \item The convergence of the encoder is slower due to the absence of a direct steering mechanism other than the decoder because of vanishing gradients.
\end{itemize}
Besides, according to the ablation studies in \cite{qi2020deep, palsson2019convolutional, ranasinghe2020convolutional},
\begin{itemize}
    \item The convolutional layers merely distil abstract spatial patterns of the \hsi\ into the abundance maps rather than learning the association amongst the pixels.
    \item There is no forcing mechanism to support the spatial correlation in abundance values since the reference for the loss function is the original \hsi.
\end{itemize}
Further, with the end-to-end architecture of AE for blind unmixing,
\begin{itemize}
\item The decoder will be limited to a single-layer network with linear activations to ensure the resemblance to matrix decomposition \cite{8322133,9342727}.
    \item The non-linearity in spectral information is modelled \cite{9423808, 9432042, zhao2021lstm}, rather than a non-linear mixing process.
\end{itemize}

% In this regard, we propose the three-network \hu\ architecture (\arch). 
In this regard, we propose the three-network \hu\ architecture (\arch). The proposed architecture is trained through two major stages. In the initial stage, the architecture referred to as ‘Guided Encoder-Decoder Architecture for Hyperspectral Unmixing with Spatial Smoothness’ (\seq) is followed. In the second stage, two variants \comb\ and \sig\ are introduced by altering the main architecture (\seq). 

In GAUSS, the first two networks: the ‘approximation network’ (AN) and ‘unmixing network’ (\hu), are encoders, and the ‘mixing network’ (MN) is the final network that serves as the decoder of the \arch. The AN is introduced to improve the association between the centre pixel and its neighbourhood in order to improve contextual spatial information in the abundance map. The output of the AN is compared against the actual spectral signature of the centre pixel to ensure the convergence of the network. The estimated centre pixel is used as the input to the UN. Furthermore, it is used as the output reference to compute the reconstruction error for the optimization of the MN.

Next, we investigate using a pseudo-ground truth, created by first segmenting the HSI with k-means and then converting it to a one-hot encoded representation for abundances to optimize the unmixing process of the spectral signatures. The pseudo-ground truth provides gradients for the UN to optimize its weights, untied from the decoder gradients. This opposes the conventional optimization of the unmixing process of AEs, where the unmixing optimization is conditioned on the HS pixel or image reconstruction. Hence, the pseudo-ground truth steers the output of the UN towards the latent space spanned by the actual abundances.  

After initializing through the \garc\, in the second phase, we train the network introducing several changes to the architecture resulting in two variants with improved performance. First, inspired by \cite{palsson2020convolutional}, the untied MN will be merged with the UN in the \comb\ method, while only allowing the encoder’s training to optimize abundances. Further, merging the UN and the MN in the second stage of training allowed the exclusion of the pseudo-ground truth. Then, in the second variant, \sig, the original \garc\ is kept intact while being retrained with a more refined pseudo ground truth generated by a specifically selected SP algorithm \cite{qian2010l1} based on its unmixing performance. 

The decoder of a typical AE architecture used in existing methods for HU is single-layered. The introduction of pseudo ground truth to train the UN prior to the training of the MN adds more flexibility to the decoder enabling the addition of more layers to it. The use of a pseudo-ground truth instils the spatial context and basic texture of the abundance map for the UN. This training of the UN prior to MN ensures the generated abundance to be at least an approximation abundance for the corresponding pixel. The input to the MN is the output provided by the UN while its’ output is the reconstructed pixel. Hence, the functionality of the decoder is to find the layer weights which will eventually generate the spectral signature once the abundances are provided. Therefore, because of this explicit nature of the input in the proposed architectures, the decoder needs not to be confined to a single layer. Thereby, with the design flexibility of the decoder, it is not bounded to perform a merely linear unmixing. Furthermore, it provides the possibility of unmixing following a linear or a non-linear premise.

In summary, the primary contribution of this article is twofold.
% \begin{enumerate}
%     \item Introduction of a novel architecture enabling preservation of the smoothness and spatial correlation of the local subspace structure and improving the neighbouring pixels' association with the centre pixel.
%     \item Introduction of a pseudo-ground truth at the end of an encoder decouples the dependency of the abundance map from the decoder optimization process, enabling the unmixing network to find the solution space independently.
%     \item Introduction of the pseudo-ground truth mechanism enables greater flexibility for the decoder structure and multiple opportunities for subsequent abundance generation steering through refined pseudo-ground truths.
%     \item While expediting the convergence, the guided latent feature space extraction process directs the network's learning process towards an optimum solution. It restrains the solution space and enables the injection of a more refined pseudo-ground truth at subsequent training rounds for further refinement or steering of the unmixing solution.
%     \item This architecture can model the non-linearity of the underlying mixing process as multiple non-linear layers can be stacked to form the decoder.
% \end{enumerate}

\begin{enumerate}
    \item Introduction of a novel architecture enabling preservation of the smoothness and spatial correlation of the local subspace structure and improving the neighboring pixels’ association with the centre pixel.
    \vspace{0.1in}
    
    Centre pixel estimation through the neighborhood has enabled the injection of spectral correlation and smoothness of the given local environment to be inhibited into the unmixing process.
    This generative process ensures the functional continuity among the pixels thus ensuring spectral smoothness and continuity while subjected to the constraints of the given environment.
    Hence, it has the added advantage of imposing a degree of smoothness that is inherent to the given environment.
    \vspace{0.1in}

    \item  Introduction of a pseudo-ground truth at the end of the UN decouples the encoder abundance map generation process and the decoder mixing for end member generation allowing the encoder and decoder to be optimized separately.
    \begin{enumerate}
        \item Decoupling the dependency of abundance map generation of encoder from the decoder optimization enables the unmixing network to find the solution space separately. While the addition of a pseudo ground truth steers the learning process of the encoder towards an optimum solution, through successive training rounds more refined pseudo-ground truth can be injected to further refine the unmixing solution.
        
        \item Improved flexibility of the decoder releases the single layer constraint which binds the performance of the decoder to be linear, therefore allowing the possibility of modeling the underlying non-linearity of the mixing process.
        
    \end{enumerate}
    
    \item A more complex and novel synthetic data set has been created for comparison purposes which incorporates fast varying irregularities, and pure regions to emulate the environment of a real \hsi. The process of generation of the synthetic data set is explained in section IV - B.
    
\end{enumerate}

Further expanding on the capabilities of the GAUSS initializer, we proposed two variants: \comb\ and \sig, with modifications to the architecture in the former and to the pseudo-ground truth in the latter based upon the flexibility introduced by the initial training stage. The introduction of the two variants established a framework to interpret variances in unmixing performances of different HU algorithms for various HS datasets and a simulated dataset with reasonable spatial complexity. The remainder of the paper is as follows. In \sref{section: problem formulation}, the concept of the GAUSS initializer and the variants are delineated followed by \sref{section: methodology} with implementation details for the HUA. Finally, the experiments and the result comparison with competing algorithms are presented in \sref{section: experiments and discussions} with explanations for the exceptional performance of the two variants ascribing to the flexibility offered by GAUSS and the effective combination of different strategies for a multitude of datasets.

% Further, expanding on the capabilities of the original \seq\ algorithm, we proposed two variants: \comb\ and \sig, with modifications to the architecture in the former and to the pseudo-ground truth in the latter as a result of the flexibility allowed by the original architecture. Also, the introduction of the three variants established a framework to interpret variances in unmixing performances of different \hu\ algorithms for various \hs\ datasets. The remainder of the paper is as follows. In \sref{section: problem formulation}, the concept of the \seq\ is delineated followed by \sref{section: methodology} with implementation details for the \arch. Finally, the experiments and the result comparison with state-of-the-art algorithms are presented in \sref{section: experiments and discussions} with explanations for the exceptional performance of the two variants ascribing to the flexibility offered by \seq\ and effective combination of different strategies for a multitude of datasets. 

\section{Problem formulation and the principles of the solution}
\label{section: problem formulation}

\begin{figure}
\centering
\includegraphics[width=0.5\columnwidth]{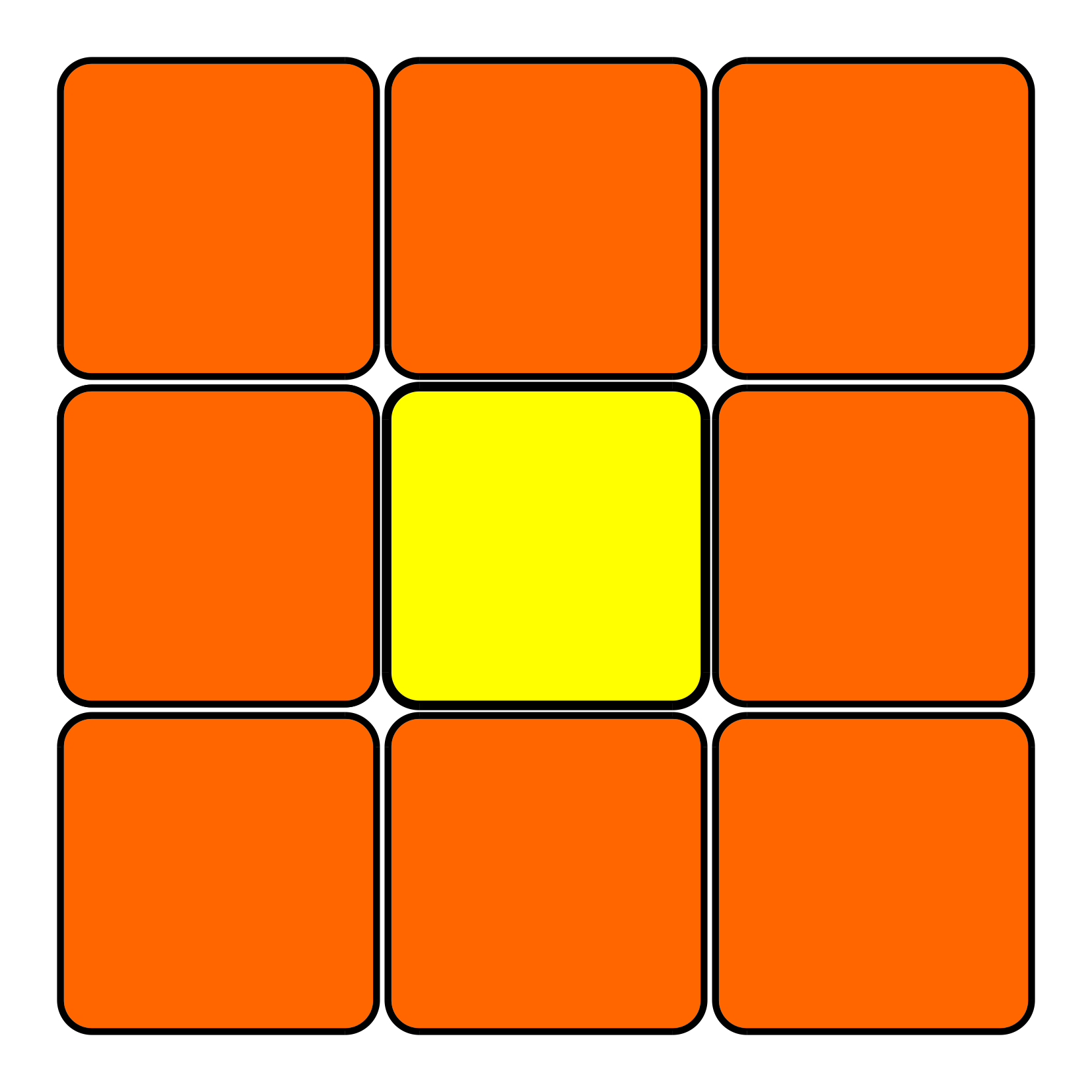}
\caption{Moore neighbourhood graph for the approximation network where the centre pixel is in yellow and its neighbourhood in orange}
\label{figure: neighbourhood graph}
\end{figure}

\begin{figure*}
    \centering
    \includegraphics[width=\textwidth]{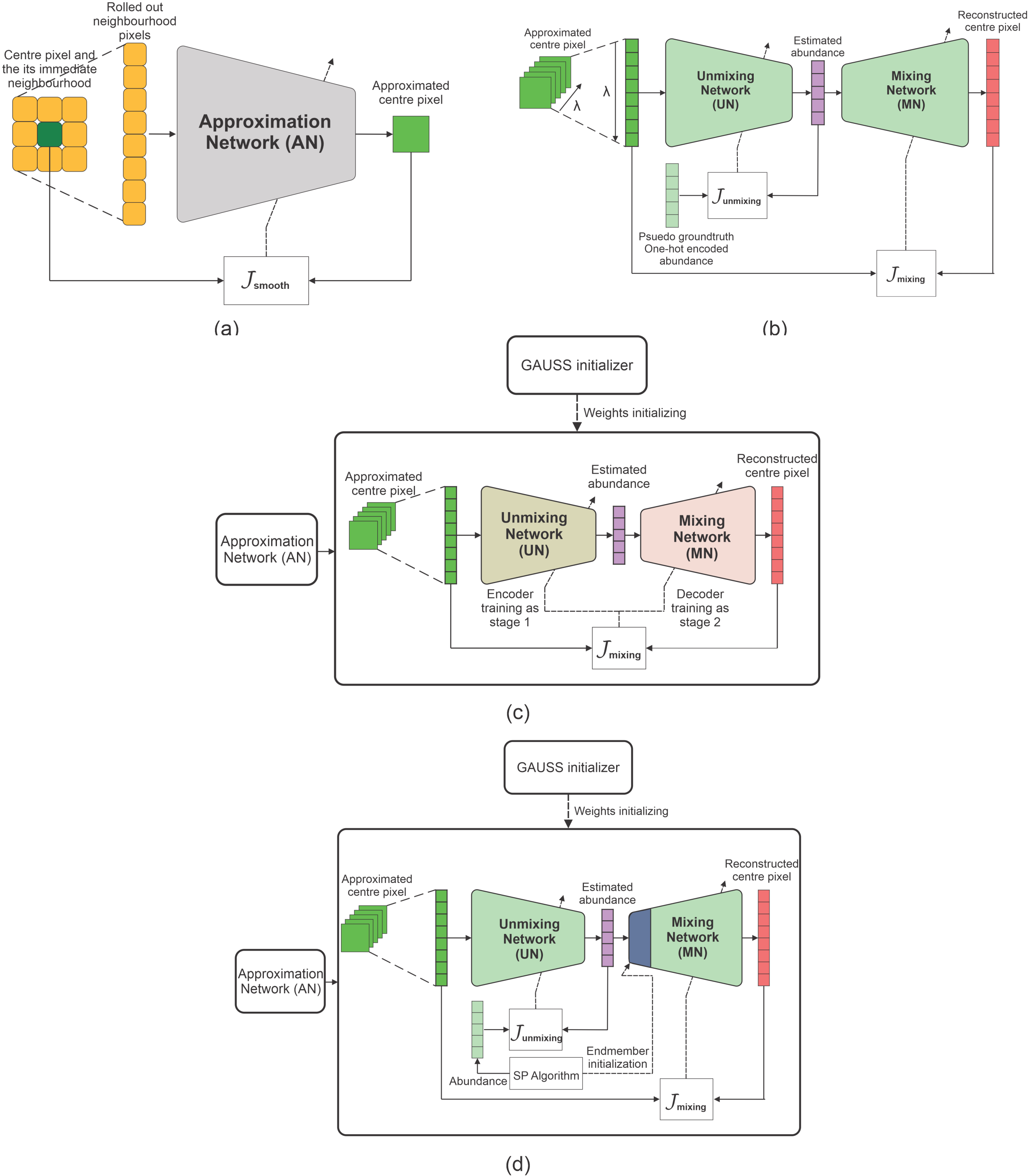}
    \caption{ (a) \textbf{\seq\ Initializer} - AN training to associate neighborhood spatial information with center pixels. (b) \textbf{\seq\ Initializer} - Sequentially trained; UN with pseudo ground truth and MN with reconstruction loss. (c) \textbf{\comb\ } - Sequentially trains UN then MN using the same reconstruction loss. (d) \textbf{\sig\ } -  Sequentially trained; UN with signal processing based abundances and MN using reconstruction loss. 
    }
    \label{figure: model architecture}
\end{figure*}

\subsection{Notation}
\label{ssection: notation}
Scalars are denoted by both lower (\scalar{x}) and uppercase (\scalar{X}) normal fonts. Boldface lowercase fonts (\mat{x}) are for 1-D arrays, and 1-D arrays are ordered as column vectors. Then, boldface uppercase fonts (\mat{X}) are for both 2-D and 3-D arrays.

Consider a \hsi, \mat{H} $\in \mathbb{R}_{+}^{w \times h \times B}$, with a width and a height of \scalar{w} and \scalar{h} in pixels, respectively, and \scalar{B} spectral bands. The \hs\ datacube can be rearranged as a matrix of size $B \times N$, where \scalar{N} is the total number of observations (pixels) and is the product of \scalar{w} and \scalar{h}. Accordingly, the \hsi\ can be denoted as \mat{Y} $= \big[ \mathbf{y}_1~\cdots~\mathbf{y}_j~\cdots~\mathbf{y}_N \big] \in \mathbb{R}_{+}^{B \times N}$ and each column $\mathbf{y}_j \in \mathbb{R}_{+}^B$ is the spectrum of the $j$\textsuperscript{th} pixel. The spectral signature of the pixel at the (\scalar{m},\scalar{n}) location of \mat{H} corresponds to the \scalar{j}\textsuperscript{th} column vector of \mat{Y}, such that $j = \{w(m-1) + n ~\vert~1 < m \leq h,~1 < n \leq w\}$ where \scalar{m} and \scalar{n} is the row and column position of the pixel, respectively.

\subsection{Mixture Models}
\label{ssection: mixture models}
The linear mixture model (\lmm) depends on the assumption that the incident light is reflected from the macroscopic surface only once and is measured by the sensing device of the imaging spectrometer without scattering. For its simplicity, \lmm\ is the most frequently used model for \hu\ in the literature \cite{heylen2014review}. The luxury of the \lmm\ is that a linear combination of endmember spectra can be written for each observation as %below,

\begin{equation}
    \label{equation: linear mixture model}
    \mathbf{y}_j = \sum_{k=1}^{K}\mathbf{S}_{kj}\mathbf{a}_k + \mathbf{e}_j
\end{equation}
where $\mathbf{S}_{kj}$ is the fractional composition of the $k$\textsuperscript{th} endmember in the $j$\textsuperscript{th}, $\mathbf{a}_k \in \mathbb{R}_{+}^{B}$ is the spectrum of the $k$\textsuperscript{th} endmember of the \hsi, $\mathbf{e}_j \in \mathbb{R}_{+}^{B}$ is an additive disruption due to noise and modeling errors, and \scalar{K} is the predetermined number of endmembers in the \hsi. The elements in $\mathbf{y}_j$ and $\mathbf{a}_k$ are non-negative as the \hsi\ is represented using the reflectance values because it inherits atmospheric corrections and sensor independencies. Due to physical implications, the abundance non-negativity constraint (ANC) $\mathbf{S}_{kj} \geq 0$ and the abundance sum-to-one constraint (ASC) $\sum_{k=1}^{K} \mathbf{S}_{kj} = 1$ are imposed in \eref{equation: linear mixture model} to assure that the fractional compositions are non-negative and the \hsi\ only consists of the predetermined endmembers. Under the \lmm, the \hsi\ can be represented using the matrix notation as

\begin{equation}
    \label{equation: LMM matrix form}
    \mathbf{Y} = \mathbf{A} \times \mathbf{S} + \mathbf{E}
\end{equation}
where $\mathbf{A} \in \mathbb{R}_{+}^{B \times K}$ is the endmember matrix whose columns are the spectra of each of the \scalar{K} endmembers, $\mathbf{S} \in \mathbb{R}_{+}^{K \times N}$ is the abundance matrix with the fractional composition of each of the $N$ pixels is organized as a column vector, and $\mathbf{E} \in \mathbb{R}^{B \times N}$ is the noise matrix. Accordingly, the extraction of the endmember spectra (\mat{A}) and their abundances (\mat{S}) are the functions of \hu.

When the assumptions of the LMM cannot be satisfied, a non-linear interaction between the signatures will occur. Several models have been introduced to capture the non-linear behaviour of the mixing process \cite{dobigeon2016linear}. Intimate mixtures exploit the fact that various particles are in very close vicinity; hence, non-linear interactions between the constituents at microscopic and macroscopic levels. Due to the multipath and scattering, interactions of the photons from different sources give rise to non-linearity. Bi-linear mixture models such non-linear interactions considering that order of interactions is two. Moreover, polynomial mixture models consider the within component interactions as well. 

\subsection{Neighbourhood approximation}
\label{ssection: neighbourhood approximation}
While the endmember spectra contain information about the spectral response of the underlying materials in the \hsi, the spatial attributes of that image are stored in the abundance matrix. Therefore, in literature, to increase spatial smoothness and spatial correlation of the fractional composition, several regularization functions have been introduced in the objective function \cite{smooth2, smooth1, he2017total, sun2018constraint, yang2015geometric, rathnayake2020graph, zhou2020subspace} to optimize \hu. However, given that the spectral signature of a given \hsi\ is the product of the endmember spectra and its fractional abundances, the spatial correlation and smoothness could be increased using the spectral signatures of the neighbouring pixels.

If the centre pixel can be approximated from its immediate neighbourhood, then the spatial correlation and smoothness shall be established in spectral signatures, thereby in abundance values. Consider the pixel at the (\scalar{m},\scalar{n}) location of the \hsi\ (\mat{H}) and the set ($\mathcal{S}$) of immediate neighbourhood pixels as illustrated in \fref{figure: neighbourhood graph}. The set can be defined as, $ \mathcal{S} = \big\{\mathbf{H}_{i,j}~|~\mathbf{H}_{i,j} \in \mathbb{R}_{+}^{B}, i \in \{m-1, m+1\},~j \in \{n-1, n+1\},~i,j \in \mathbb{Z}^{+}\big\}$ and the cardinality of the set is equal to the number of neighbourhood pixels considered during the construction of the set and it is denoted by $\vert \mathcal{S}\vert $.

Then following the pixel to column position transition explained in \sref{ssection: notation}, by rearranging the vectors in $\mathcal{S}$ as a single column vector denoted by $\mathbf{n}_j$, the following relationship can be formulated between the centre pixel ($\mathbf{y}_{j}$) and its neighbourhood. 

\begin{equation}
    \centering
    \label{equation: neighbourhood connection}
    \mathbf{y}_{j} \approx \mathbf{W_{n}}\mathbf{n}_{j}
\end{equation}
where $\mathbf{W_n} \in \mathbb{R}^{B \times B\vert \mathcal{S}\vert}$ approximates the centre pixel using its neighbourhood pixel vector $\mathbf{n}_{j}$. To learn the spatial correlation across the \hsi, $\mathbf{W_n}$ can be trained using the following objective function,

\begin{equation}
    \centering
    \label{equation: spatial smoothness}
    \argmin_{\mathbf{W_n}} {\Vert \mathbf{Y} - \mathbf{W_n}\,\mathbf{N} \Vert}^{2}_{F}
\end{equation}
Here, \mat{N} is the matrix constructed by concatenating the neighbour vectors of each centre pixel, and \mat{Y} is the matrix form of the \hsi\ and $\Vert \cdot \Vert_{F}$ is the Frobenius norm. Then by performing \hu\ on pixels generated from the trained $\mathbf{W_n}$ in (\ref{equation: spatial smoothness}), the spatial correlation and smoothness could be brought forward to fractional abundances by the AN displayed in \fref{figure: model architecture}.

\subsection{Controlled abundance estimation}
\label{ssection: controlled abundance estimation}
Generally, \dl\ unmixing is performed using the AE architecture with the pixel reconstruction error as the objective function to optimize. However, a drawback of using the AE for unmixing is that it precludes direct optimization of abundances in contrast to non-\dl\ methods. Moreover, a conventional AE architecture may suffer from the vanishing gradient problem as the network becomes deeper and deeper. That is because gradients descending through back-propagation may exponentially decrease when they reach the layers in the encoder. In addition, the unmixing process is dependent on the decoder performance of the AE because of the back-to-back connection between the encoder and the decoder. 

Consequently, to avoid the dependency on the endmember signature optimization from the abundance optimization, the proposed work considers an additional objective function to train the network for abundance optimization. We create a reference for the abundance maps to train the UN, first by segmenting the \hsi\ with k-means and converting the result to its one-hot encoded representation. This reference for the abundances is referred to as the pseudo-ground truth and is described in \sref{ssection: pseudo-ground truth preparation for abundances}. Further, the injection of a pseudo-ground truth facilitated by the split architecture provides a mechanism to reinforce the abundance optimization. It allows for a more refined pseudo-ground truth from unsupervised \sgp\ algorithms as proposed in the \sig\ variant (see \sref{ssection: alternative training strategies}) for subsequent training for better accuracy.

Each pixel's fractional abundances can be considered a probability distribution with the ASC and ANC. Each abundance value for the corresponding endmember is the individual probability. This definition allows us to reconsider the unmixing process as a classification using categorical-cross entropy as the objective function. When the UN learns hidden representations of the \hsi\ by mapping the spectral signatures onto the probability space using the pseudo-ground truth, pixels that are not comprised of pure pixels, \textit{i.e.} mixed pixels, will be forced towards a particular endmember. However, since the UN is a transformation on the pixel signature and the signature of a mixed pixel is different from a pure signature, the fractional abundances can be driven towards pure abundances without losing the generalization of the UN. As these two objectives behave as opposite forces, the fractional abundances will be at an equilibrium rather than driven by a single cost function.

\subsection{Expanding decoder flexibility}
\label{ssection: expanding decoder flexibility}

AEs for \hu\ are omnipresent because they resemble the matrix decomposition used in blind source separation. Since the encoder task is designed to transform the pixel or image of the \hs\ dataset into a latent representation, there is no limitation on the number of layers in the encoder. However, unlike the encoder, the decoder of the AE does not emulate a process; instead performs the matrix multiplication between the abundances and endmembers. Therefore, the endmembers of the \hs\ data are extracted by the weights of the decoder. Hence, the decoder architecture is bound to a single layer decoder to ensure the output of the encoder will be the abundances.

By adopting the pseudo-ground truth mechanism, the UN is trained to produce abundance values at its output in our proposed architecture independent of the decoder architectural form. Therefore, since it is guaranteed that the input of the MN is the abundances of the centre pixel, the single-layered decoder constraint can be released without using recurrent layers as in \cite{zhao2021lstm}. Because the output of the decoder is compared against the pixel signature, the decoder requirement is to find the optimal combination of weights that will map the abundances to spectral signatures. Further, the decoder can model the non-linear mixing process with the proposed change.

Therefore, in the \seq\ architecture, the pixel's neighbourhood is fed to the AN, and its output is compared against the actual centre pixel and the input to the UN. The UN will generate the abundances of the centre pixel at its output which is subsequently used as the input to the MN. After training, endmembers are generated by feeding an identity matrix ($\mathbf{I} \in \mathbb{R}^{K \times K}$) to the MN, which represents pure pixel feeds because, with the modifications to the decoder architecture, the endmembers can no longer be extracted from the weights of the MN.

%%\subsection{Network architecture variants}
%%\label{ssection: three algorithms introduction}
%%Using the AN and the pseudo-ground truth mechanism leads to variants in the architecture. As illustrated in \fref{figure: model architecture}, three possible architectures: \seq, \comb, and \sig\ are proposed with different alternatives for the pseudo-ground truth and the encoder-decoder training.

\section{Methodology}
\label{section: methodology}

This section elaborates the proposed architecture for \hu\ with spatial smoothness. The proposed method is primarily based on the traditional AE model used in standard \hu\ practice under \dl. However, the proposed method can be viewed as three separate networks instead of the standard AE model in \hu\ due to the method of training adapted in this work. The architecture of the aggregated network is illustrated in \fref{figure: model architecture}(a). The AN is used to impose spatial smoothness on the \hsi\ as discussed in \sref{ssection: neighbourhood approximation}. The network approximates the centre pixels $\mathbf{Y} \in \mathbb{R}_{+}^{B \times N}$ using their neighbourhood pixels $\mathbf{N} \in \mathbb{R}_{+}^{B \vert \mathcal{S}\vert \times N}$ given by

\begin{equation}
    \label{equation: smoothness network}
    \mathbf{\hat{Y}} = f_n(\mathbf{N})
\end{equation}
with $f_n \colon \mathbb{R}_{+}^{B \vert \mathcal{S}\vert \times N} \to \mathbb{R}_{+}^{B \times N}$. The network learns the parameters for approximation by minimizing the average reconstruction error between the input \mat{Y} and the reconstructed spectral signatures \mat{\hat{Y}} from the neighbourhood pixels. Therefore, after the learning process is terminated, the smoothness matrix (\mat{W_n}) in (\eref{equation: spatial smoothness}) and the spectral signatures with improved spatial characteristics can be realized by

\begin{align}
    \text{Smoothness transitions} & \colon f_n (\mathbf{N}) \Rightarrow \mathbf{W_n} \mathbf{N}\\
    \text{Smoothed pixels} & \colon \mathbf{\hat{Y}}.
\end{align}

Next, to learn the hidden representation of the \hsi, the UN is trained, which transforms the smoothed spectral signatures (\mat{\hat{Y}}) from the AN to a low dimensional representation ($\mathbf{\hat{L}} \in \mathbb{R}_{+}^{K \times N}$) given by

\begin{equation}
    \centering
    \mathbf{\hat{L}} = f_e(\mathbf{\hat{Y}}),
\end{equation}
with $f_e\colon \mathbb{R}_{+}^{B \times N} \to \mathbb{R}_{+}^{K \times N}$. The output from the UN is fed to the MN, which decompresses the hidden representation to reconstruct the smoothed spectral signatures at its output. The decoder function, therefore, can be represented as

\begin{equation}
    \centering
    \mathbf{\hat{\hat{Y}}} = f_d(\mathbf{\hat{L}})
\end{equation}
where $f_d \colon \mathbb{R}_{+}^{K \times N} \to \mathbb{R}_{+}^{B \times N}$, and $\mathbf{\hat{\hat{Y}}}$ is the output of the decoder network. After the successful termination of the learning process for \hu, the unmixing results can be extracted by$\colon$

\begin{equation}
    \label{equation: parameter extraction}
    \begin{aligned}
    \centering
    \text{Abundance estimation} &\colon \mathbf{\hat{L}}^{*} \Rightarrow \mathbf{\hat{S}}\\
    \text{Endmember estimation} &\colon f_d^{*}(\mathbf{I_{K}}) \Rightarrow \mathbf{\hat{A}}
    \end{aligned}
\end{equation}
where $\mathbf{I_K} \in \mathbb{R}_{+}^{K \times K}$ is the identity matrix and * denotes the optimal solution at the end of the training process.

\subsection{Encoder}
\label{ssection: encoder}
The encoder combines the AN and UN to improve spatial characteristics and learn hidden representations in this work. For each pixel of the \hsi, we first consider its immediate neighbourhood and reconstruct the centre pixel from the set of neighbourhood pixels. Then, the output of the AN is sent to the UN to learn the unmixing process and estimate abundances. The architecture of the encoder is given in \fref{figure: model architecture}(a), and fully connected layer (FCNL) architecture is used in constructing the encoder.

In our work, the fractional abundances are considered probabilities, and learning hidden representations of the \hsi\ is similar to performing a multi-class classification with the pseudo-ground truths. 
In our work, the encoder is designed to use spatial information and optimize the estimation of fractional abundances without interference from the decoder optimization. To optimize the AN, we use the reconstruction error of the centre pixel from its neighbourhood as given by

\begin{equation}
    \label{equation: reconstruction error for smoothing}
    \mathcal{J}_{\text{smooth}} = \frac{1}{N} \Vert \mathbf{Y} - \mathbf{\hat{Y}}\Vert_{F}^{2}
\end{equation}
where $\Vert \cdot \Vert_{F}$ is the Frobenius norm. Then, to abide by the ANC and ASC, we apply the following scaling and standardization at the final layer of the UN defined as follows,

\begin{equation}
    \label{equation: softmax activation}
    \hat{l}_{kj} = \frac{e^{-z_{kj}}}{\sum_{k=1}^{K} e^{-z_{kj}}}
\end{equation}
where \scalar{z_{kj}} is the output of the \scalar{k}\textsuperscript{th} node at the final layer of the encoder prior activation and \scalar{\hat{l}_{kj}} is the estimated fractional abundance for the $k$\textsuperscript{th} endmember of the $j$\textsuperscript{th} column in \mat{\hat{L}}.
The standardization to probability distributions given in \eref{equation: softmax activation} can be realized through \textit{Softmax} activation function of the output layer of the encoder. Then to optimize the unmixing process of the encoder, the categorical cross-entropy is used as below,

\begin{equation}
\begin{aligned}
    \label{equation: categorical cross-entropy}
    \mathcal{J}_{\text{unmixing}} = -\frac{1}{N} \sum_{\substack{j,k=1}}^{N,K} l_{kj}\log\hat{l}_{kj} + (1-l_{kj})\log(1-\hat{l}_{kj})
\end{aligned}
\end{equation}
where \scalar{l_{kj}} are abundance values for \scalar{k}\textsuperscript{th} endmember of the \scalar{j}\textsuperscript{th} column of the pseudo-ground truth used to expedite optimization of the UN.

\subsection{Decoder}
\label{ssection: decoder}
An MLP-based network is adopted for the decoder in our proposed scheme, with its structure reported in \fref{figure: model architecture}. In this work, the decoder is designed to reconstruct the input to the UN, which comprises smoothed pixels generated from the AN of the proposed architecture. As discussed in \sref{ssection: encoder}, with a pseudo-ground truth mechanism for abundance estimation, the decoder architecture is not restricted to a single-layered linear network; therefore, it can learn non-linear relationships as well.

The decoder of the proposed network learns non-linearity relationships between abundances and endmember signatures during the mixing process because of the introduced extension in the layers. Nevertheless, as mentioned in \sref{ssection: decoder}, by extending the decoder layers, the extraction of the endmember signatures cannot be done through the network weights because the unmixing process is now not limited to a single layer rather laid out over a single network with multiple layers. However, since the decoder is trained by using abundances generated from the encoder network, the decoder can reconstruct the corresponding spectral signature for any abundance vector in the low dimensional space learned by the encoder. Hence, by feeding this one-hot encoded vector to the decoder at the end of weight training, the spectral information of the corresponding endmember can be extracted. For example, to find the endmember signature of the \scalar{k}\textsuperscript{th} endmember, we will construct a one-hot vector \mat{o_k} such that $\mathbf{o}_k = \big[ 0,\cdots0,1,0\cdots,0 \big]$ where the unity entry is at the \scalar{k}\textsuperscript{th} position of the vector. The vector format implies the existence of a pure pixel containing only that end member. Hence, the resulted spectral signature of the MN is simply the corresponding endmember signature. Then by concatenating these one-hot vectors of each endmember, we can construct an identity matrix, when given to the decoder, which will output the endmember signatures of the \hsi\ in a single attempt as presented in \eref{equation: parameter extraction} for endmember estimation.

To train the decoder network, we diverge from the conventional practice of using the mean squared error between the encoder input $\mathbf{Y}$ and the decoder output of each pixel $\hat{\mathbf{Y}}$ defined as 

\begin{equation}
    \mathcal{J}_{\text{reconstruction}} = \frac{1}{NB}\big\Vert \mathbf{Y} - \mathbf{\hat{\hat{Y}}} \big\Vert_{F}^{2}
    \label{equation: pixel reconstruction error}
\end{equation}
for the \hsi. Rather, we adopt the pixel spectral information divergence (pSID) proposed by \cite{chang1999spectral} as the reconstruction error to optimize the mixing process, which is defined by,

\begin{equation}
    \label{equation: spectral information divergence}
    \begin{aligned}
    \mathcal{J}_{\text{mixing}} = &D(\mathbf{\hat{Y}} \Vert \mathbf{\hat{\hat{Y}}}) + D(\mathbf{\hat{\hat{Y}}} \Vert \mathbf{\hat{Y}})\\
     = &\frac{1}{N}\sum_{\substack{j,b=1}}^{N,B} \Big\{q_{bj} \log \Big( \frac{q_{bj}}{\hat{q}_{bj}} \Big) + \hat{q}_{bj} \log \Big( \frac{\hat{q}_{bj}}{q_{bj}} \Big)\Big\}
    \end{aligned}
\end{equation}
where
$\mathbf{q}_{j} = \sfrac{\mathbf{\hat{y}_j}}{\Vert\mathbf{\hat{y}}_j\Vert_1}$ and
$\mathbf{\hat{q}}_j = \sfrac{\mathbf{\hat{\hat{y}}}_j}{\Vert\mathbf{\hat{\hat{y}}}_j\Vert_1}$ are the probability distribution vectors of the spectral signature of the \scalar{j}\textsuperscript{th} pixel of \mat{\hat{Y}} and \mat{\hat{\hat{Y}}}, respectively. Accordingly, $\mathbf{q}_{bj}$ and $\mathbf{\hat{q}}_{bj}$ represent the probability values of the \scalar{b}\textsuperscript{th} band of the \scalar{j}\textsuperscript{th} pixel. Then, $\Vert\cdot\Vert_1$ is 1-norm of the corresponding spectral signature.
In \eref{equation: spectral information divergence}, $D(\cdot\Vert\cdot)$ is called the relative entropy between the smoothed pixels at the UN input and reconstructed pixels at the decoder output. The pSID defined by \eref{equation: spectral information divergence} can be used to measure the spectral similarity between corresponding pixel vectors in $\mathbf{\hat{Y}}$ and $\mathbf{\hat{\hat{Y}}}$.

\subsection{Pseudo-ground truth preparation for abundances}
\label{ssection: pseudo-ground truth preparation for abundances}
In the original \seq\ architecture, one-hot encoded representation is used as the pseudo-ground truth to guide the UN of the encoder. To construct the one-hot encoded representation of the \hsi, we first segment the \hsi\ using the k-means algorithm on the spectral signatures. Further, the injection of the pseudo-ground truth enables greater flexibility for the decoder structure and multiple opportunities for subsequent abundance generation steering through refined pseudo-ground truths as proposed below.

\subsection{Variants of the \seq\ algorithm}
\label{ssection: alternative training strategies}
The \seq\ architecture is the foundation that uses the one-hot encoder abundances as the pseudo-ground truth. Then, we propose two architecture variants: \comb\ and \sig, for \hu\ that are pre-trained using the \seq\ method. First, following \cite{palsson2020convolutional}, we will connect the decoder to the UN and remove the pseudo-ground truth. Then, we will train the network, similar to the conventional \dl\ unmixing method, without allowing the decoder to train. This strategy will allow the unmixing process to optimize abundance results according to pixel reconstruct error defined by \eref{equation: spectral information divergence}. After its completion, the decoder of the \arch\ will be trained while the encoder training is withheld. For reference, this training method is called `\comb' (see \fref{figure: model architecture}(b)).

Second, analogous to using unsupervised \hu\ algorithms for endmember weight initialization, we consider abundance results of a \sgp\ algorithm with reliable unmixing performance as a pseudo-ground truth for \arch. This training strategy is referred to as the `\sig' method (see \fref{figure: model architecture}(c)). The \sig\ method is a combination of both \seq\ and \comb\ strategies because and facilitates a hybrid modus operandi by fusing the best of the abilities of the other two methods. However, the output of the UN will be compared against the new pseudo-ground truth. Further, in this scheme, the encoder and the decoder will also be trained successively like in the \comb\ strategy. Nonetheless, the subsequent refinement in abundance optimization steering through the insertion of abundance maps generated via high-performing \sgp\ was enabled by using the pseudo-ground truth mechanism. Accordingly, we use the abundance results of the \elhalf\ algorithm for all the datasets to maintain consistency even though it could be any contextual driver for abundance generation.

\subsection{Implementation and training of the network}
\label{ssection: training of the unmixing network}
The \arch\ was developed in TensorFlow v2 using the Google Colaboratory using default settings for optimization parameters, and the detailed architecture of the \arch\ is given in \tref{table: network parameters}. Then the Adam optimizer was used to optimize the three networks, and for the input-output dataset preparation, we used a batch size of 32 for each \hsi\ dataset.

In all three training strategies, the encoder and the decoder are successively trained, and the section was trained for 25 epochs because the \comb\ and \sig\ methods are used with the network pre-trained by the \seq\ method. Accordingly, in both \comb\ and \sig, the \arch\ was first trained for 25 epochs under the \seq\ method, and another 25 epochs either with the \comb\ and \sig\ method totalled 50 epochs. Since the total number of epochs for the \seq\ method is different from that for the other two methods, the pre-trained AE was further trained for additional 25 epochs with the \seq\ method to enable fair comparison amongst the proposed three algorithms. 

The pseudo-ground truth used for the \sig\ method was the abundance results from the \elhalf\ algorithm considering its abundance performance for all three real datasets.

% RGB images of the dataset
\begin{figure}[!b]
\centering
\begin{tikzpicture}
    \node (img1)
    {\includegraphics[scale=0.23, angle=0]{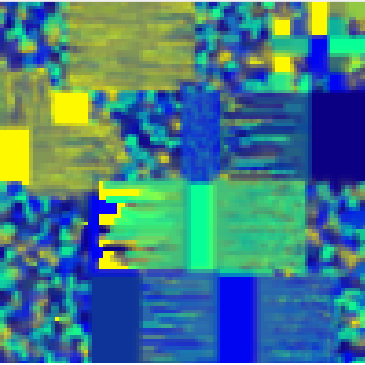}};
    \node[below of = img1, anchor=center, yshift=-0.05cm, font=\mapfont{5}\small] {(a)};
    \node[right=of img1,xshift=-1.2cm] (img2) 
    {\includegraphics[scale=0.45]{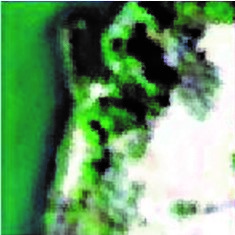}};
    \node[below of = img2, anchor=center, yshift=-0.05cm, font=\mapfont{5}\small] {(b)};
    \node[right=of img2,xshift=-1.2cm] (img3) 
    {\includegraphics[scale=0.45]{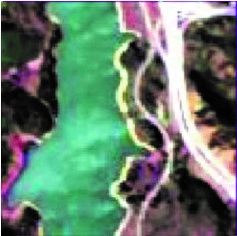}};
    \node[below of = img3, anchor=center, yshift=-0.05cm, font=\mapfont{5}\small] {(c)};
    \node[right=of img3,xshift=-1.2cm] (img4) 
    {\includegraphics[scale=0.45]{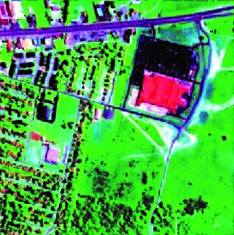}};
    \node[below of = img4, anchor=center, yshift=-0.05cm, font=\mapfont{5}\small] {(d)};

\end{tikzpicture}
\caption{RGB images of the datasets (a). Simulated (b). Samson (c). Jasper-Ridge (d). Urban}
\label{figure: RGB images}
\end{figure}

\section{Experiments and Discussions}
\label{section: experiments and discussions}

\subsection{Performance criteria}
\label{ssection: performance criteria}
The performance of the abundance estimation is measured by the average root mean square error (aRMSE), average abundance information divergence (aAID), and average abundance angle distance (aAAD). Then, the accuracy of endmember estimation from the \hu\ algorithm is evaluated using the average spectral angle distance (aSAD) and the average spectral information divergence (aSID), which are defined as follows,

\begin{align}
    % average root mean square error
    \text{aRMSE} & = \sqrt{\frac{1}{NK} \big\Vert \mathbf{S}-\mathbf{\hat{S}} \big\Vert_{F}^{2}}\\[5pt]
    % average abundance angle distance
    \text{aAAD} & = \frac{1}{N} \sum_{i=j}^{N} \mathrm{cos}^{-1}\Big( \frac{\langle\mathbf{s}_j,\mathbf{\hat{s}}_j\rangle}{\Vert \mathbf{s}_j \Vert \Vert\mathbf{\hat{s}}_j \Vert} \Big)\\[5pt]   
    % average abundance information divergence
    \text{aAID} & = \frac{1}{N}\sum_{\substack{j=1\\k=1}}^{N,K} \Big\{s_{kj} \log \Big( \frac{s_{kj}}{\hat{s}_{kj}} \Big) + 
    \hat{s}_{kj} \log \Big( \frac{\hat{s}_{kj}}{s_{kj}} \Big)\Big\}\\[5pt]
    % average spectral angle distance
    \text{aSAD} & = \frac{1}{K} \sum_{k=1}^{K} \mathrm{cos}^{-1}\Big( \frac{\langle\mathbf{a}_k,\mathbf{\hat{a_k}}\rangle}{\Vert \mathbf{a}_k \Vert \Vert\mathbf{\hat{a}}_k \Vert} \Big)\\[5pt]
    % average spectral information divergence
    \text{aSID} & = \frac{1}{K}\sum^{K,B}_{\substack{k=1\\b=1}}\Big\{p_{bk} \log \Big( \frac{p_{bk}}{\hat{p}_{bk}} \Big) +
    \hat{p}_{bk} \log \Big( \frac{\hat{p}_{bk}}{p_{bk}} \Big)\Big\}
\end{align}
where \mat{S} and \mat{\hat{S}} represents the ground truth and estimated abundances. Then, $\mathbf{s}_j$ and $\mathbf{\hat{s}}_j$ are the ground truth and estimated abundances at \scalar{j}\textsuperscript{th} column or pixel, and \scalar{s_{kj}} and \scalar{\hat{s}_{kj}} are the ground truth and estimated abundance value for \scalar{k}\textsuperscript{th} endmember of the \scalar{j}\textsuperscript{th} pixel. Next, $\mathbf{a}_k$ and $\mathbf{\hat{a}}_k$ are ground truth and extracted endmember for the \scalar{k}\textsuperscript{th} source, and \scalar{p_{bk}} and \scalar{\hat{p}_{bk}} are probability value for the \scalar{b}\textsuperscript{th} spectral band of the \scalar{k}\textsuperscript{th} ground truth and extracted endmembers, respectively. The probability vector for the \scalar{k}\textsuperscript{th} endmember is computed as $\mathbf{p}_k = \sfrac{\mathbf{a}_k}{\Vert \mathbf{a}_k \Vert}$ and $\mathbf{\hat{p}}_k = \sfrac{\mathbf{\hat{a}}_k}{\Vert \mathbf{\hat{a}}_k \Vert}$ for the ground truth and extracted signatures, respectively.

% Network parameters
\begin{table}[!t]
\centering
\captionsetup{justification=centering, labelsep=newline}
\caption{Network parameters of the autoencoder for unmixing}
\footnotesize
\resizebox{\columnwidth}{!}{
\begin{tabular}{c c c c c}
\hline \hline
\\[-2ex]
\multicolumn{2}{c}{Network}	& Layer description	& Output shape	& Bias state	\\[1ex]
\hline
\\[-1ex]
\multirow{14}{*}{\rotatebox[origin=c]{90}{Encoder}}	&\multirow{6}{*}{\rotatebox[origin=c]{90}{\begin{tabular}{c} Approximation\\[-0.8ex] network \end{tabular}}}	&Input layer	&[(None, $\vert \mathcal{S} \vert \times B$)]	&N/A	\\[2ex]
	&	&Dense layer 1	&(None, $\lfloor\vert\mathcal{S}\vert\times\sfrac{B}{2}\rfloor$)	&TRUE	\\[2ex]
	&	&Dense layer 2	&(None, $\lfloor\vert\mathcal{S}\vert\times\sfrac{B}{8}\rfloor$)	&TRUE	\\[2ex]
	&	&Dense layer 3	&(None, $B$)	&TRUE	\\[4ex]
&\multirow{6}{*}{\rotatebox[origin=c]{90}{\begin{tabular}{c}Unmixing \\[-0.8ex] network \end{tabular}}}	&Input layer	&[(None,$B$)]	&N/A	\\[2ex]
	&	&Dense layer 1	&(None, $\lfloor\sfrac{B}{2}\rfloor$)	&FALSE	\\[2ex]
	&	&Dense layer 2	&(None, $\lfloor\sfrac{B}{8}\rfloor$)	&FALSE	\\[2ex]
	&	&Dense layer 3	&(None, $K$)	&FALSE	\\[4ex]
\multirow{4}{*}{\rotatebox[origin=c]{90}{Decoder}} &	&Input layer	&[(None, $K$)]	&N/A	\\[2ex]
	&	&Dense layer 1	&(None, $\lfloor\sfrac{B}{4}\rfloor$)	&FALSE	\\[2ex]
	&	&Dense layer 2	&(None, $B$)	&FALSE	\\
\hline \hline
\\[-1.5ex]
\multicolumn{5}{l}{\small $\lfloor\cdot\rfloor$ represents the floor operation.}
\end{tabular}}
\label{table: network parameters}
\end{table}

\subsection{Experimental setting}
\label{ssection: experimental setting}

To evaluate the performance of \seq in a complex environment an algorithm that is capable of generating simulated datasets with a sufficient amount of complexity was designed. Compared with the simulated data sets in the literature, the dimensions were selected to be $100\times100$ with a spectral resolution of 198 bands.  

For the generation of abundances, initially, a zeros tensor of size $100\times100\times4$  was segmented into sixteen $25\times25\times4$ tensors. Next, the algorithm randomly selects one from the segmented tensors and initiates the assignment of abundance fractions. This is done through a superpixel of size  $3\times3$. Thereafter, a random selection of a center pixel is done and checked for occupancy. If the pixel is not already occupied then the possibility of it being on the outer edge is investigated. If the pixel is not a part of the outer edge then the superpixel of the above-mentioned dimension is constructed and the abundance fractions are assigned while satisfying the sum to one constraint.If the selected center pixel is at an edge,then necessary alterations are done to the superpixel to get adapted according to the nature of the edge. However, following this procedure will not make the dataset realistic which contains pure regions and sparse sections as remote sense HSIs. To achieve this, some of the $25\times25$ were allotted to be pure or to be to maintain a constant abundance fraction variation. This process makes sharp edges. To resolve this issue, initially, the sharp edges were extended irregularly, and then the entire image was smoothened through mean filtering. The generated image is shown in \fref{figure: RGB images}.

To generate the synthetic image next endmember assignment was carried out. For this, initially, four endmembers were chosen from the United States Geological Survey (USGS)\footnote{\url{https://www.usgs.gov/labs/spectroscopy-lab/science/spectral-library}} endmember library \cite{kokaly2017usgs} was utilized. The spectral signatures of "Ilmenite, Montmorillonite, Limestone, and Tree" were chosen as the candidate endmembers. Then, as the spectral signatures extracted contained bands which were not calibrated preprocessing was carried out for the removal of not calibrated bands. Furthermore, as the spectral signatures of different elements were acquired from different sensors there was a slight deviation in the wavelengths present in each spectral signatures. To overcome this issue cubic interpolation was used to generate spectral signature containing the same wavelengths.  

Moreover, a set of three real \hs\ datasets\footnote{\url{http://lesun.weebly.com/hyperspectral-data-set.html}}: Samson, Jasper-Ridge, and Urban were used for quantitative evaluation purposes. A precise ground truth for the endmembers and abundances for each dataset is available on each website. The Samson dataset consists of three endmembers with 156 spectral bands covering 401 nm to 889 nm wavelengths, while Jasper-Ridge comprises four sources with 198 spectral bands ranging from 380 nm to 2500 nm. The urban dataset has 162 spectral bands spanning the region from 400 nm to 2500 nm and five endmembers. The spatial resolutions of the datasets, in units of pixels $\times$ pixels, are 95$\times$95, 100$\times$100 and 307$\times$307, respectively.  

In order to compare the results of the proposed algorithm, we have chosen eight different classical \hu\ algorithms. In addition to DL architectures such as CNNAEU \cite{palsson2020convolutional}, DAEN \cite{su2019daen}, \cite{8322133}, uDAS \cite{qu2018udas} and EGU-Net\cite{9444141}, renowned signal processing algorithms which leverage different NMF and other geometric frameworks namely, L\textsubscript{$\sfrac{1}{2}$} NMF \cite{qian2010l1}, R-CoNMF \cite{li2016robust}, MLNMF \cite{rajabi2014spectral}, SULoRA \cite{hong2018sulora} and VCA-FCLS \cite{nascimento2005vertex} were used for this purpose. The proposed and comparison algorithms were tested on the real hyperspectral datasets as well as the simulated dataset incorporating the optimum parameters suggested by the authors in their work. The RGB images of the datasets are illustrated in \fref{figure: RGB images}. 
% Tabulated results for the all datasets
\begin{table*}[t]
\centering
\captionsetup{justification=centering, labelsep=newline}
\caption{Unmixing performance comparison for the four datasets: Simulated, Samson, Jasper-ridge, and Urban. The best-performing algorithms are ranked up to the third, and the ranks are superscript.}
\footnotesize
\resizebox{\textwidth}{!}{
% \begin{tabular*}{\textwidth}{@{\extracolsep{\fill}}*{15}{c}}
\begin{tabular}{c l :c c :c c c c c :c c c c c}
\hline \hline
& &\multicolumn{2}{:c}{\textbf{Proposed methods}} &\multicolumn{5}{:c}{\textbf{Deep learning methods}} &\multicolumn{5}{:c}{\textbf{Signal processing methods}}\\
% \cline{3-5} \cline{6-10} \cline{11-15}\\[-1.5ex]
&& \comb\ & \sig\  & \cnnaeu & \daen & \daeu & \egu & \udas & \begin{tabular}{c}$L_{\sfrac{1}{2}}$\\[-0.2ex] NMF \end{tabular} & R-CoNMF & MLNMF  & SULoRA & \begin{tabular}{c}VCA\\[-0.2ex] FCLS \end{tabular}\\
\hline
\multirow{5}{*}{\rotatebox[origin=c]{90}{\textbf{Simulated}}} &aRMSE	&0.2725	&0.1312\sups{1}	&0.2453	&0.2550	&0.2371\sups{3}	&0.2783	&0.2349	&0.2419	&0.1618\sups{2}	&0.2896	&0.3724	&0.2874
\\
&aAAD	&0.6077	&0.3414\sups{1}	&0.8532	&0.7386	&0.8297	&0.6844	&0.8430	&0.6909\sups{3}	&0.4256\sups{2}	&0.8240	&0.7732	&0.8168
\\
&aAID	&1.7655	&0.2496 &3.4171	&1.5346	&2.8457	&2.3981\sups{3}	&7.1713	&2.5311	&0.9826\sups{2}	&3.3080	&3.4135	&3.9947	\\

&aSAD	&0.1325	&0.2459	&0.1316	&0.0301	&0.0756	&0.1059	&0.3111	&0.0008\sups{1}	&0.0269\sups{3}	&0.0984	&0.5475	&0.0009\sups{2}	\\
&aSID	&0.1874\sups{2}	&0.2739	&0.3764	&0.5461	&0.3246	&0.2332	&0.4308	&0.3436	&0.0003\sups{1}	&0.2467\sups{3}	&1.2544	&0.3433	\\[1.5ex]
\hdashline\\[-1ex]
\multirow{5}{*}{\rotatebox[origin=c]{90}{\textbf{Samson}}} & aRMSE	&0.1348\sups{2}	&0.0467\sups{1}	&0.3724	        &0.3909	&0.3902	&0.1573\sups{3}	&0.3027	&0.2689	        &0.3330	&0.3424	&0.3553	&0.3996	\\
&aAAD	&0.3899	&0.0724\sups{1}	&0.9073	        &0.7774	&0.8384	        &0.2518\sups{2}	&0.5793	&0.4991\sups{3}	&1.0979	&0.6533	&0.7054	&0.8825	\\
&aAID	&5.7561	&4.4736\sups{3}	&7.6865	        &6.7775	&5.5561\sups{2}	&1.0455\sups{1}	&4.7794	&4.0999	\sups{2}        &10.7096	&6.4476	&6.4119	&8.0004	\\
&aSAD	&0.0345\sups{2}	 &0.0211\sups{1}	&0.1769	        &0.1413	&0.1155	&0.0603\sups{3}	&0.0697	&0.0811	        &0.2451	&0.0897	&0.0990	&0.1364	\\
&aSID	&0.1326	        &0.0584\sups{3}	&1.5040	&1.5217	&2.4887	&0.0486\sups{2}	&0.0328\sups{1}	&0.1953	        &0.4725	&0.4758	&0.3954	&0.6746	\\[1.5ex]
\hdashline\\[-1ex]
\multirow{5}{*}{\rotatebox[origin=c]{90}{\textbf{Jasper-ridge}}} &aRMSE	&0.1224\sups{3}	&0.0483\sups{1}	&0.2888	&0.2227	&0.2623	&0.4651	&0.1190\sups{2}	&0.3393	&0.3570	&0.4135	&0.3494	&0.4903	\\
&aAAD	&0.6732	&0.1489\sups{1}	&0.7186	&0.5001\sups{3}	&0.6239	&1.2312	&0.2177\sups{2}	&0.8437	&1.0048	&1.0811	&0.8596	&1.2592	\\
&aAID	&12.8246	&12.4485	&6.8717\sups{3}	&4.5497\sups{2}	&7.4464	&15.9190	&1.5840\sups{1}	&9.6208	&8.6153	&11.3338	&10.9363\sups{3}	&19.8816	\\
&aSAD	&0.0513\sups{1}&0.0965\sups{2}	&0.1552	&0.1328	&0.1634	&0.1447	&0.1062	&0.1677	&0.1422\sups{3}	&0.2241	&0.1893	&0.2092	\\
&aSID	&0.2126	&0.1487\sups{2}	&0.2635	&0.2092\sups{3}	&0.2710	&1.3000	&0.0441\sups{1}	&0.8390	&0.2914	&1.1527	&0.8440	&1.3342	\\[1.5ex]
\hdashline\\[-1ex]
\multirow{5}{*}{\rotatebox[origin=c]{90}{\textbf{Urban}}}&aRMSE	&0.0912\sups{1}	&0.1921\sups{2}	&0.3203	&0.2454	&0.3777	&0.5097	&0.3425	&0.4231	&0.3330	&0.4135	&0.3187	&0.2061\sups{3}	\\
&aAAD	&0.7321\sups{2}	&2.9351	&0.9947  &0.7583\sups{3} &1.0989	&1.4478	&1.0362	&1.2346	&1.0979	&1.0811	&0.9746	&0.6786\sups{1}	\\
&aAID	&14.8739	&11.5265	&10.2948\sups{3}	&7.9027\sups{1}	&12.2434	&27.6442	&14.9865	&20.8305	&10.7096	&11.3338	&14.2694	&7.9499\sups{2}	\\
&aSAD	&0.0043\sups{1}	&0.5814	&0.2070	&0.1753	&0.1878	&0.1351	&0.0158\sups{2}	&0.1258	&0.2451	&0.2241	&0.1141	&0.0956\sups{3}	\\
&aSID	&0.0231\sups{1}	&0.4354\sups{3}	&0.9303	&1.0081	&1.2864	&1.3433	&0.7105	&0.9319	&0.4725	&1.1527	&0.5245	&0.1208\sups{2}	\\
\hline \hline
% \end{tabular*}
\end{tabular}
}
\label{table: unmixing performance}
\end{table*}

\subsection{Experiments on simulated data}
\label{ssection: experiments on simulated data}

When comparing the performance of existing algorithms, signal processing-based approaches perform better than \dl\ unmixing methods. As a whole compared to other \dl\ methods the proposed methods have superior performance. 

The conventional unmixing algorithms used in the analysis are geometry-based methods, and their derivation is based on ideal \hsi s. The simulated dataset resonates with the requirement of those algorithms.
The presence of pure or marginally-pure pixels favouring unmixing algorithms that exploit underlying geometric frameworks to prevail over \dl\ unmixing algorithms is reflected through these results.

However, the results of \sig\ method are pre-eminent in abundance estimation. Notably, the estimated abundances of \sig\ have managed to be better than its pseudo-ground truth abundances from the \elhalf\ algorithm. This result can be taken as evidence for the fact that the supervisory inputs used in the \sig\ method have contrived the abundance estimation to be accurate even for a complex environment.

With regard to end member extraction, SP based \elhalf\ and RCo-NMF provide best results for aSAD and aSID error matrices respectively while the \comb\ has achieved the second finest result for aSID.  

%%%%%%%%%%%%%%%%%%%%%%%%%%%%%%%%%%%%%%%%%%%%%%%%%%%%%%%%%%
%sir

% Abundance figures for all datasets
\begin{figure*}
    \centering
    \includegraphics[width=0.8\textwidth]{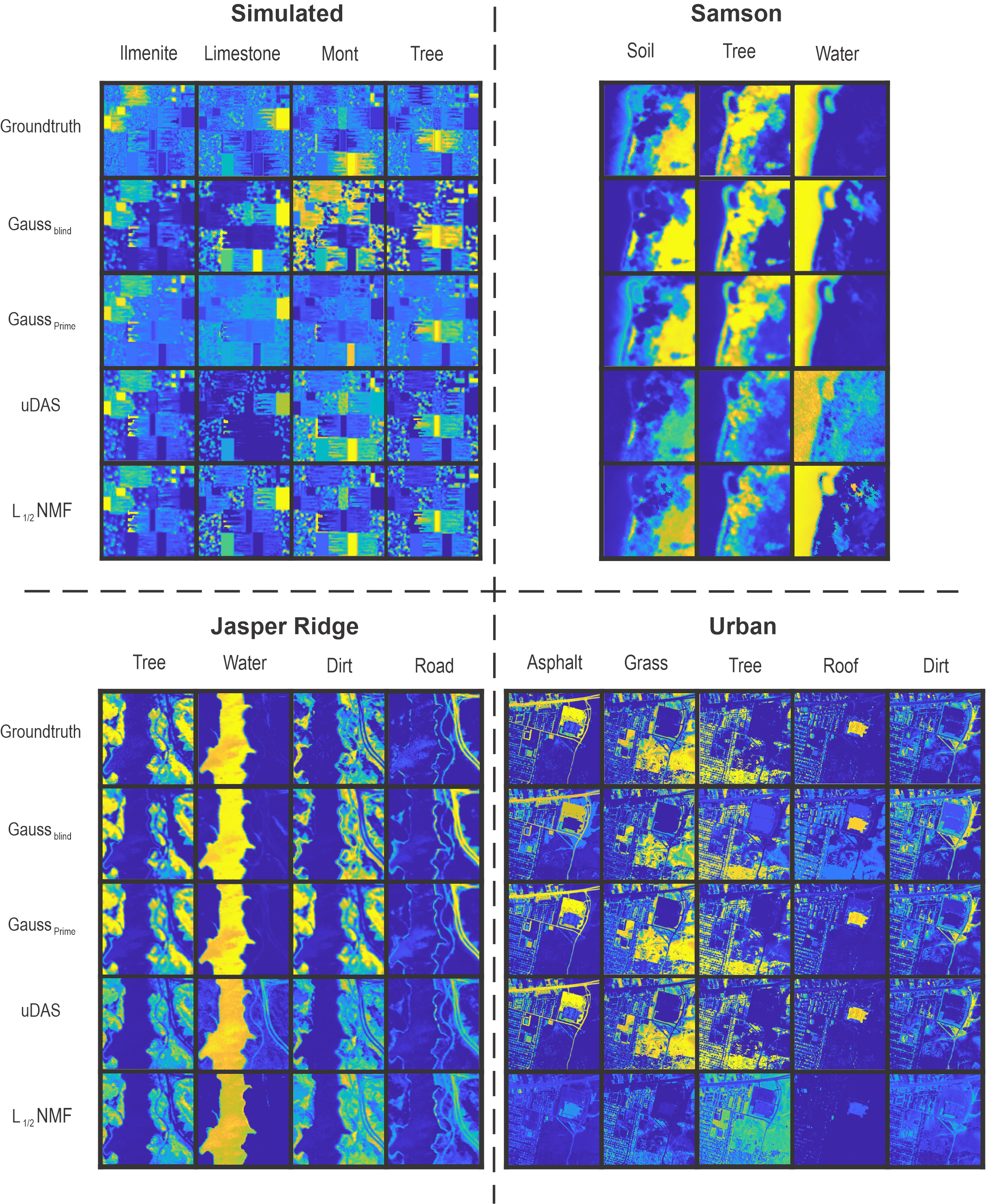}
    \caption{Groundtruth and abundance maps for Simulated, Samson, Jasper Ridge and Urban data sets from \sig\, \comb\, uDAS and \elhalf\ }
    \label{figure: abundance results}
\end{figure*}

% Endmember signature 

\begin{figure*}
    \centering
    \includegraphics[width=0.98\textwidth]{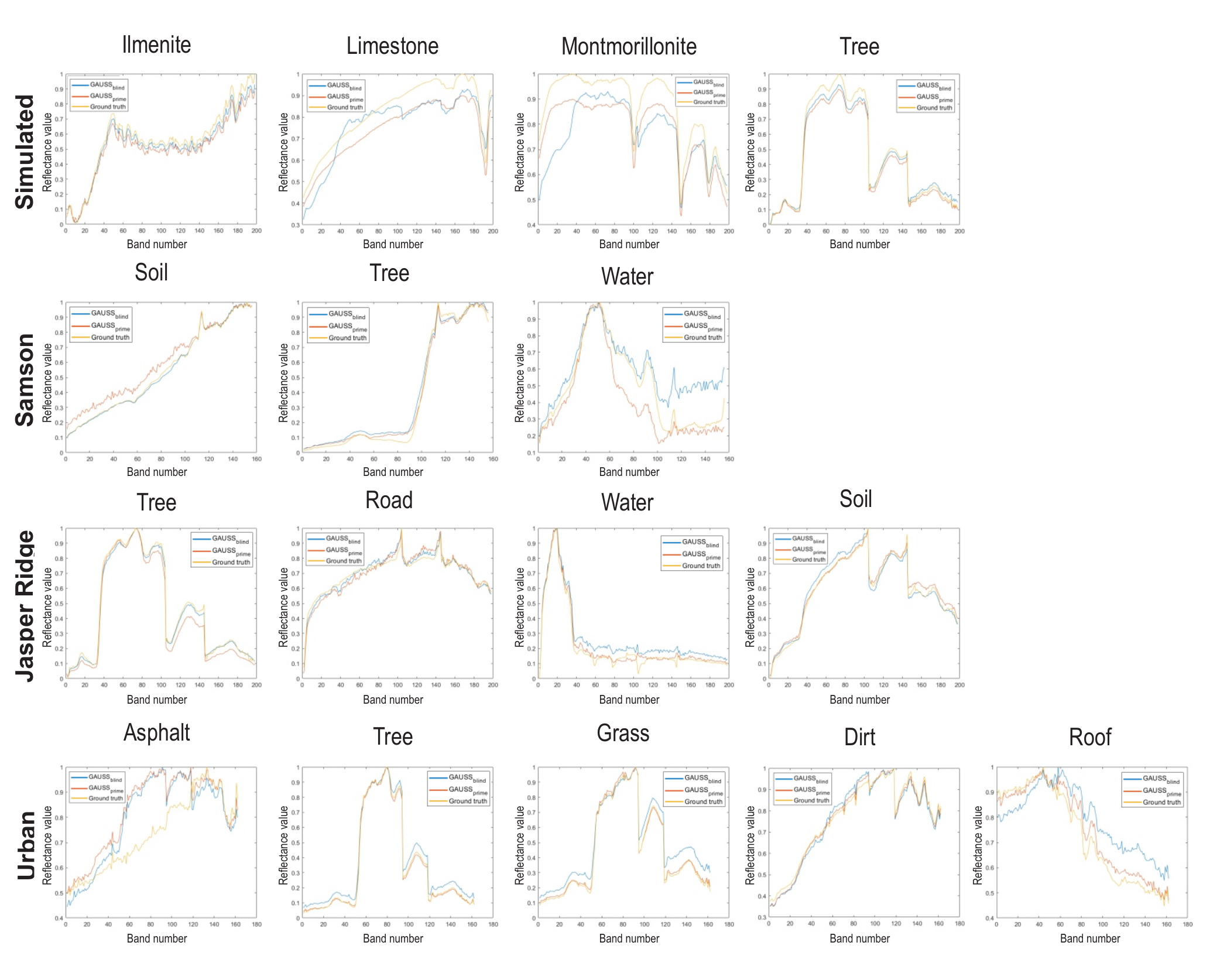}
    \caption{Real endmember and signatures extracted for Simulated, Samson, Jasper Ridge and Urban data sets from \sig\, \comb\, uDAS and \elhalf\ }
    \label{figure: endmember results}
\end{figure*}

%%%%%%%%%%%%%%%%%%%%%%%%%%%%%%%%%%%%%%%%%%%%%%%%%%%%%%%%%%

\begin{table}[!t]
\centering
\captionsetup{justification=centering, labelsep=newline}
\caption{ Result variances after 5 consecutive trials}
\footnotesize
\resizebox{0.75\columnwidth}{!}{
% \begin{tabular*}{\textwidth}{@{\extracolsep{\fill}}*{15}{c}}
\begin{tabular}{c l c c}
\hline \hline
& & \sig\ & \comb \\
\hline
\multirow{5}{*}{\rotatebox[origin=c]{90}{\textbf{Simulated}}} &aRMSE	&4.3*10\sups{-5} &1.6*10\sups{-4}
\\
&aAAD	&2.1*10\sups{-4} &7.2*10\sups{-5}
\\
&aAID	&9.6*10\sups{-5} &2.5*10\sups{-5} 	
\\
&aSAD	&6.4*10\sups{-5} &1.4*10\sups{-6}		
\\
&aSID	&7.3*10\sups{-4} &1.7*10\sups{-5}		
\\[1.5ex]
\hdashline\\[-1ex]
\multirow{5}{*}{\rotatebox[origin=c]{90}{\textbf{Samson}}} &aRMSE	 &6.35*10\sups{-5} &4.46*40\sups{-7}\\
&aAAD	&2*10\sups{-3} &5*10\sups{-3}	
\\
&aAID	&6.23*10\sups{-5} &8.22*10\sups{-5}
\\
&aSAD	&8.07*10\sups{-5} &2.26*10\sups{-5}	
\\
&aSID	&5*10\sups{-5} &7*10\sups{-4}
\\[1.5ex]
\hdashline\\[-1ex]
\multirow{5}{*}{\rotatebox[origin=c]{90}{\textbf{Jasper-ridge}}} &aRMSE	&7.9*10\sups{-5} &3*10\sups{-4}
\\
&aAAD	&2*10\sups{-4} &1*10\sups{-3}
\\
&aAID	&4.2*10\sups{-5} &6.3*10\sups{-5}	
\\
&aSAD	&1.2*10\sups{-4} &3*10\sups{-4}	
\\
&aSID	&2.1*10\sups{-4} &6.9*10\sups{-3}
\\[1.5ex]
\hdashline\\[-1ex]
\multirow{5}{*}{\rotatebox[origin=c]{90}{\textbf{Urban}}}&aRMSE	&2.3*10\sups{-4} &3.9*10\sups{-4}	\\
&aAAD	&1.3*10\sups{-2} &2.8*10\sups{-2}
\\
&aAID	&5.3*10\sups{-3} &4.7*10\sups{-3}		
\\
&aSAD	&7.8*10\sups{-3} &8*10\sups{-6}
\\
&aSID	&1.6*10\sups{-2} &9.2*10\sups{-5}	\\
\hline \hline
% \end{tabular*}
\end{tabular}
}
\label{table: variance}
\end{table}

\subsection{Experiments on real data}

\subsubsection{Samson dataset}

It is imperative to understand the Samson dataset to analyze the performance of the two methods. It has three principal endmembers, and the boundaries of these endmembers are almost non-overlapping. 
The performance of the \comb\ and the \sig\ variants is contributed by the one-hot encoded pseudo ground truth in the initial stage of the training. Similarly, by the \elhalf\ pseudo ground truth in the \sig\ variant and the combined training of the encoder and decoder in the \comb\ variant during the latter stages. Consequently, the pre-training of the HUA under the \seq\ architecture has driven the optimization towards an optimal solution space that further optimizes the reconstruction error gradients.

Next, the \sig\ has performed the best in terms of abundance estimation with the exception of AID. Moreover, using the \elhalf\ abundances as the pseudo-ground truth, the \sig\ method has produced results better than the \elhalf\ algorithm. The reason is that the network had benefited from pre-training the network using the \seq\ model and \elhalf\ method, which has learned different properties of the latent space of the HSI. Further, the fact that \sig\ continues to outperform \elhalf\ is a clear evidence of optimal combining, and that it is better than the \comb\ is because reconstruction optimization is not as effective as \elhalf\ steering due to the specific landscape. 

%%As the smaller endmember count and the sparsity present in the samson dataset result a simple geometry within the latent space which benefit the geometry-based SP algorithms, the cumulative error matrices show better performance in SP than DL based algorithms mostly. This supports the decision taken to introduce abundances from a conventional unsupervised algorithm as the pseudo-ground truth in \sig\ method. 

While \sig\ stands on top with regard to SAD, the performance of uDAS is higher when considering SID. Furthermore, the endmember performance of the \sig\ method has better results than that of the \elhalf\ algorithm, which produced the pseudo-ground truths for the \sig\ method. Therefore, as the \sig\ method has outperformed \elhalf\ algorithm in both abundance and endmember performances, the performance increment should be from combining the best attributes of the \seq\ initializer and the \elhalf\. The injection of conventional SP as a supervisory input has managed 
to generate a  significant performance boost in \hu\ compared to pure \dl\ methods. This validates the combining of the two domains which has been done in the \sig\ architecture which is made possible by the introduction of the supervisory ground truth for UN training by the \garc\. Consequently, the decoder of the \sig\ method has benefited from using the AN in the HUA and the pseudo-ground truth from the \elhalf\ algorithm. Since the AN improves the association between the neighboring pixels and the centre pixel, the sparsity in the HSI resulted from the direct impact of the \elhalf\ algorithm will be complemented. As the output of the encoder is fed to the decoder and its output is compared against the pixels that promote sparsity in the image, the decoder benefits from the properties used by other unmixing algorithms. This combination of different aspects of algorithms via dedicated networks signifies the use of the AN and a pseudo-ground truth, as proposed in this work.

\subsubsection{Jasper-ridge dataset}

The Jasper-ridge dataset has four endmembers, and their boundaries are not as non-overlapping as in the Samson dataset.  

\sig\ has been able to achieve the best results for abundance estimation except for the aAID error matric. The much superior ground truth in the latter stages and the added effectiveness of the AN’s spatial smoothing mechanic might have helped the proposed \comb\ and  \sig\ to perform better than the other DL methods. Here, the pseudo-ground truth from the \elhalf\ abundances is considered superior to the more rudimentary one-hot encoded maps used in conditioning the network through \seq\ initializer. The \seq\ initializer input information on pure or primarily pure abundances while the \elhalf\ algorithm has information about abundances with lower fractions. Therefore, using the two pseudo-ground truths in consecutive training stages has complemented each other in the \sig\ method, which reaffirms the benefits of using supervisory data to train the encoder, initially proposed for the first training stage (\seq\ architecture) with the k-means algorithm.

In the Jasper-ridge dataset, the road endmember is challenging to extract since it is scarce compared to other endmembers and heavily correlates with the dirt endmember, despite having high reflectance values. As a result, most algorithms have not successfully extracted the road endmember, especially conventional algorithms and ultimately its abundance values. However, according to \fref{figure: endmember results}, all two methods in the proposed work have extracted a valid signature for the road while \comb\, \sig\ and uDAS algorithm have had the highest success. This achievement of \seq\ variants was made possible by the AN and the pseudo-ground truth. Because the AN has managed to decorrelate the pure endmembers to conform to their high correlation, and as a result, the decoder has been allowed to extract the road endmember accurately as opposed to the comparative algorithms.

\subsubsection{Urban dataset}

For the urban dataset, the \comb\ method has produced the best results for estimating the abundance and endmembers. Even though DAEN and VCA-FLCS have good performance in terms of aAID, \comb\ provides the best results for the rest matrices. Since the \seq\ variants perform unmixing pixelwise, the benefits of the spatial features in the urban HSI should be extracted by AN, which feeds its output to the UN. 
 Also, the abundance maps for \elhalf\ in Fig. 4 have much lower abundance values for most of the endmembers, and the maps have significantly deviated from the ground truths. Since there is a marked difference between the ground truth and the \elhalf\ abundances, the UN  in \sig\ has not been able to improve using a conventional algorithm with inferior performance like with the Jasper-ridge dataset. Nonetheless, it should be noted that the pseudo-ground truth of the \sig\`  method could be any reliable SP algorithm, and the \elhalf\ algorithm was used to maintain consistency among the datasets. 
 
The \comb\ method has the best performance for endmember extraction. Therefore, it can be said that the decoder for the \comb\ method has already been initiated around the decoder’s optimal solution after the initial training stage following the \seq\ structure. According to the aSAD metric, the endmember performance is as improved as the abundance error metric. Hence, the proposed AN and the one-hot encoded pseudo-ground truth have been conducive to endmember estimation.  

According to \tref{table: unmixing performance}, the method with the least reconstruction error for each dataset has the best performing figures for unmixing results in \tref{table: unmixing performance}. Therefore, though the authors have chosen the \elhalf\ algorithm as the conventional algorithm to construct a pseudo-ground truth for the training of the UN, the selection of a suitable unsupervised algorithm can also be based on the reconstruction error of the \hsi\ for that algorithm. However, in this work, the \elhalf\ algorithm was used to maintain consistency in the comparison and remove the effect of using different algorithms in constructing the pseudo-ground truth.

\subsection{Summary}

Even though the performance of existing \sgp\ and \dl\ is competitive, the proposed \sig\ and \comb\ is comparatively superior as a whole when considering the overall results. 

For sparse environments like Samson and Jasper Ridge, the performance of \sig\ is considerably high compared to all other methods. This might be due to the presence of many pure pixels which complement the usage of k-means in the stage of training and \elhalf\ in the latter stages. Further explanation can be found in section III - D.

% However, the performance of EGU-Net for the Samson dataset and uDAS for the Jasper Ridge dataset cannot be disregarded.

While considering environments that are complex with sudden and scattered variations of abundances like the Urban dataset, the results have scattered among \dl\ and \sgp\ domains equally. However, in terms of the urban dataset, the performance of the gauss blind method is noteworthy. 

% When considering the simulated dataset, though the performance of a particular algorithm cannot be distinguished as the best performer, in terms of abundance estimation the results of the \sig\ depict an improved performance.

The consistency of performance of the introduced algorithms is assured through summarizing the results for five consecutive trials in \tref{table: unmixing performance}. The variances of the obtained results are illustrated in \tref{table: variance}

\begin{table}[!t]
\centering
\captionsetup{justification=centering, labelsep=newline}
\caption{Ablation study for the \comb\ method}
\footnotesize
\resizebox{0.97\columnwidth}{!}{

\begin{tabular}{c l c c c c c}
\hline \hline
\\[-1.5ex]
& & aRMSE & aAAD & aAID & aSAD & aSID 
\\ 
\hline
\\[-1.5ex]
\multirow{6}{*}{\rotatebox[origin=c]{90}{\textbf{Samson}}} & AE	 & 0.4150 & 0.7826 & 6.4259 & 0.1416 & 1.5271\\
& AN + AE & 0.4386 & 0.7470 & 6.1648 & 0.1113 & 1.2928 \\
& AN + pGT & 0.4030 & 0.7241 & 6.2504 & 0.1140 & 0.0634 \\
& AE + AN + pGT & 0.2150 & 0.4350 & 5.9902 & 0.0947 & 0.0725 \\
& AE + AN + GAUSS & 0.1348 & 0.3899 & 5.7561 & 0.0345 & 0.1326 \\
& initializer + pGT & & & & & 
\\[1ex]
\hdashline
\\[-1ex]

\multirow{6}{*}{\rotatebox[origin=c]{90}{\textbf{Urban}}}& AE	& 0.1793 & 0.7823 & 15.7651 & 0.0217 & 0.0288	\\
& AN + AE & 0.1158 & 0.8328 & 15.1476 & 0.0145 & 0.0286 \\
& AN + pGT & 0.1073 & 0.8876 & 15.0938 & 0.0074 & 0.3674 \\
& AE + AN + pGT & 0.0988 & 0.8743 & 15.0076 & 0.0106 & 0.0302 \\
& AE + AN + GAUSS & 0.0912 & 0.7321 & 14.8739 & 0.0043 & 0.0231 \\
& initializer + pGT & & & & & 
\\[2ex]
\hline \hline
% \end{tabular*}
\end{tabular}
}
\label{table: ablation Gblind}
\end{table}

%%%%
\begin{table}[!t]
\centering
\captionsetup{justification=centering, labelsep=newline}
\caption{Ablation study for the \sig\ method}
\footnotesize
\resizebox{0.97\columnwidth}{!}{

\begin{tabular}{c l c c c c c}
\hline \hline
\\[-1.5ex]
& & aRMSE & aAAD & aAID & aSAD & aSID 
\\ 
\hline
\\[-1.5ex]
\multirow{4}{*}{\rotatebox[origin=c]{90}{\textbf{Samson}}} & AE + pGT	& 0.4186 & 0.6975 & 5.4374 & 0.0857 & 0.8212\\
& AN + AE + pGT & 0.1989 & 0.6183 & 4.1830 & 0.0218 & 0.0657 \\
& AE + AN + GAUSS & 0.0467 & 0.0724 & 4.4736 & 0.0211 & 0.0584 \\
& initializer + pGT & & & & & 
\\[1ex]
\hdashline
\\[-1ex]

\multirow{4}{*}{\rotatebox[origin=c]{90}{\textbf{Urban}}}& AE + pGT	& 0.3426 & 3.0756 & 11.7028 & 0.7006 & 0.5257	\\
& AN + AE + pGT & 0.2342 & 2.9588 & 11.9285 & 0.6283 & 0.4769 \\
& AE + AN + GAUSS & 0.1921 & 2.9351 & 11.5265 & 0.5814 & 0.4354 \\
& initializer + pGT & & & & & 
\\[2ex]
\hline \hline
% \end{tabular*}
\end{tabular}
}
\label{table: ablation Gprime}
\end{table}

Here, AE refers to the AutoEncoder, AN refers to the Approximation Network, and pGT refers to the Pseudo ground truth. 

\section{Ablation study}
\label{section: ablation Gprime}

An ablation study has been conducted to further illustrate the effectiveness of \comb\ and \sig\ architectures on \hu. Samson and Urban data sets were used for this study, and the findings are tabulated in \tref{table: ablation Gblind} and \tref{table: ablation Gprime}. Furthermore, It is evident that all quantitative error metrics show the effectiveness of the two variants.

\section{Conclusion}
\label{section:Conclusion}

In this paper, we proposed two algorithms for \hu. In this work, we discuss split training of the encoder-decoder with an additional network (AN) preceding the UN and the use of a pseudo-ground truth for abundances for supervised learning of the UN. Then, the AN approximates centre pixels from their neighborhood to improve the association between nearby pixels and spatial correlation; and performs decorrelation of pixel and endmember impurities. In addition, the use of the pseudo-ground truth accelerated the encoder training. By releasing the single-layer constraint on the decoder it enables the possibility of modeling the underlying non-linearity of the mixing process in remote sensing and enabling subsequent training procedures with more refined abundance maps as pseudo-ground truths for improved performance. Using the pseudo-ground truth created with k-means segmentation resulted in abundance binarization, but it was conducive to estimating pure endmembers of the \hsi within the initial training stage.

Based on the initial training architecture \seq\ two variants were established. In the \comb, we combine the encoder and decoder to back-propagate the reconstruction error gradients to both the decoder and the encoder. Then, in the \sig\, we experimented with using the abundance results of a \sgp-based unmixing method as the pseudo-ground truth with the combined encoder-decoder architecture, which was made possible due to the introduced pseudo-ground truth mechanic that decoupled and enabled such abundance steering. Furthermore, the use of \elhalf\ allows the pathway to further expedite on using more refined pseudo-ground truth from the \sgp\ and \dl\ domains. 

The proposed variants mostly outperformed or produced results competitive with existing unmixing algorithms from \dl\ and \sgp\ domains. In comparison with the existing \dl\ and \sgp\ approaches for \hu\  the performance of the proposed models can be distinguished for improved performance in terms of the considered error matrices. Besides, we discussed the effect of the dataset on the performance of the unmixing method and the selection of the unmixing algorithm out of the proposed methods for an unseen hyperspectral dataset. Meanwhile, the same technique could be applied to deciding on an algorithm to create the pseudo-ground truth.

% References
\bibliographystyle{IEEEtran}
\bibliography{IEEEabrv,reference}

% Author biographies
\vskip -30pt plus -1fil

\begin{IEEEbiography}[{\includegraphics[width=1in,height=1.25in,clip]{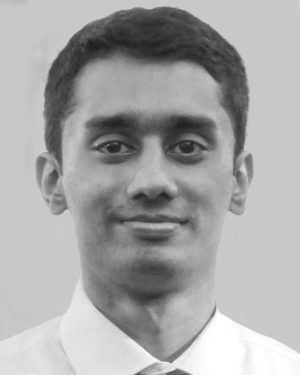}}]{D.Y.L. Ranasinghe} received the B.Sc. (Eng.) degree in electrical and electronic engineering from the University of Peradeniya, Sri Lanka, in 2020. Immediately after that, he joined the School of Engineering, Sri Lanka Technological Campus, Padukka, Sri Lanka, as a Research Assistant. He is currently working as a Research Assistant with the University of Peradeniya under a research grant from the International Development Research Centre (IDRC), Canada. He has published in IEEE JSTARS, IEEE Access, and numerous IEEE conferences. His research interests include hyperspectral and multispectral imaging, remote sensing, signal and image processing, and deep learning.
\end{IEEEbiography}

\vskip -30pt plus -1fil

\begin{IEEEbiography}[{\includegraphics[width=1in,height=1.25in,clip]{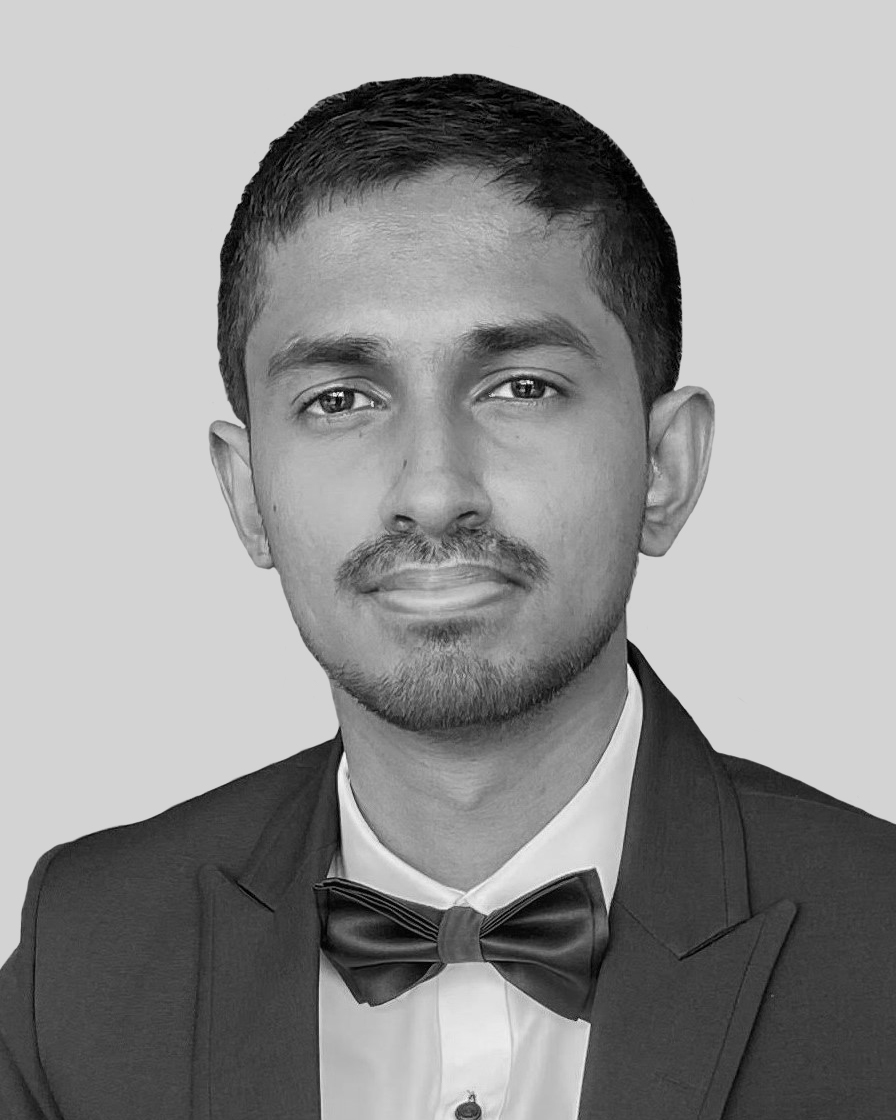}}]{H.M.H.K. Weerasooriya} received the B.Sc. degree in electrical and electronic engineering with a first-class honors from the University of Peradeniya, Peradeniya, Sri Lanka in 2020. He is currently an Instructor with the Department of Electronic and Electrical Engineering, University of Peradeniya. He is presently involved in the researches on hyperspectral imaging for remote sensing and agriculture applications, and he has numerous publications in IEEE conferences. His research interests include image processing, signal processing, communication, machine learning and deep learning.
\end{IEEEbiography}

\vskip -30pt plus -1fil

\begin{IEEEbiography}[{\includegraphics[width=1in,height=1.25in,clip]{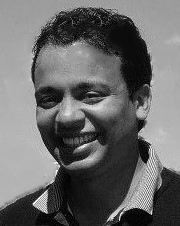}}]{G.M.R.I. Godaliyadda} (Senior Member, IEEE) obtained his B.Sc. Engineering degree in Electrical and Electronic Engineering from the University of Peradeniya, Sri Lanka, in 2005, and Ph.D. from the National University of Singapore in 2011. Currently, he is attached to the University of Peradeniya, Faculty of Engineering, Department of Electrical and Electronic Engineering as a Senior Lecturer. His current research interests include image and signal processing, pattern recognition, computer vision, machine learning, smart grid, bio-medical and remote sensing applications and algorithms. He is a Senior Member of the IEEE. He is a recipient of the Sri Lanka President's Award for Scientific Publications for 2018 and 2019. He is the recipient of multiple grants through the National Science Foundation (NSF) for research activities. His previous works have been published in IEEE-TGRS and several other IEEE-GRSS conferences including WHISPERS and IGARSS. He also has numerous publications in many other IEEE transactions, Elsevier and IET journals and is the recipient of multiple best paper awards from international conferences for his work.
\end{IEEEbiography}

\vskip -30pt plus -1fil

\begin{IEEEbiography}[{\includegraphics[width=1in,height=1.25in,clip]{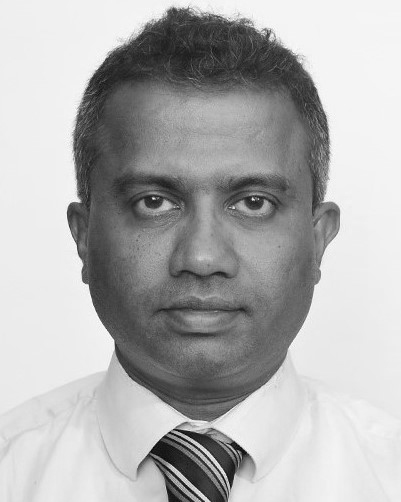}}]{H.M.V.R. Herath} (Senior Member, IEEE) received the B.Sc.Eng. degree in electrical and electronic engineering with 1st class honours from the University of Peradeniya, Peradeniya, Sri Lanka, in 1998, M.Sc. degree in electrical and computer engineering with the award of academic merit from the University of Miami, USA in 2002, and Ph.D. degree in electrical engineering from the University of Paderborn, Germany in 2009. In 2009, he joined the Department of Electrical and Electronic Engineering, University of Peradeniya, as a Senior Lecturer. His current research interests include hyperspectral imaging for remote sensing, multispectral imaging for food quality assessment, Coherent optical communications and integrated electronics. Prof.  Herath was a member of one of the teams that for the first time successfully demonstrated coherent optical transmission with QPSK and polarization multiplexing. He is a member of the Institution of Engineers, Sri Lanka and The Optical Society. He is a Senior Member of the IEEE. He was the General Chair of the IEEE International Conference on Industrial and Information Systems (ICIIS) 2013 held in Kandy, Sri Lanka. His previous works have been published in IEEE-TGRS and several other IEEE-GRSS conferences including WHISPERS and IGARSS. He received the paper award in the ICTer 2017 conference held in Colombo Sri Lanka. Prof.  Herath is a recipient of Sri Lanka President's Award for scientific research in 2013.
\end{IEEEbiography}

\vskip -30pt plus -1fil

\begin{IEEEbiography}[{\includegraphics[width=1in,height=1.25in,clip]{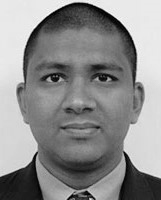}}]{M.P.B. Ekanayake} (Senior Member, IEEE) received his B.Sc. Engineering degree in Electrical and Electronic Engineering from University of Peradeniya, Sri Lanka, in 2006, and Ph.D. from Texas Tech University in 2011. Currently, he is attached to the University of Peradeniya as a Senior Lecturer. His current research interests include applications of signal processing and system modeling in remote sensing, hyperspectral imaging, and smart grid. He is a Senior Member of the IEEE. He is a recipient of the Sri Lanka President's Award for Scientific Publications in 2018 and 2019. He has obtained several grants through the National Science Foundation (NSF) for research projects. His previous works have been published in IEEE-TGRS and several other IEEE-GRSS conferences including WHISPERS and IGARSS. He also has multiple publications in many IEEE transactions, Elsevier and IET journals and has been awarded several best paper awards in international conferences.
\end{IEEEbiography}

\vskip -30pt plus -1fil

\begin{IEEEbiography}[{\includegraphics[width=1in,height=1.25in,clip]{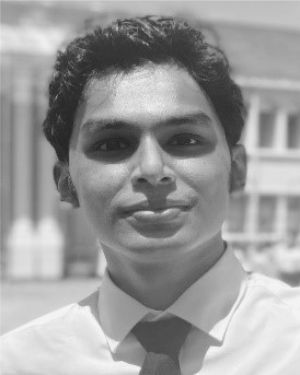}}]{D. Jayasundara} received a first-class honors degree in Electrical and Electronic Engineering from University of Peradeniya (2021). He’s currently working as an instructor at Department of Engineering Mathematics. His research interests include Mathematics, Deep Learning, Digital Communication and Signal Processing. At present, he is involved in research associated with Hyper-Spectral Imaging for Remote sensing and Applications of Multi-Spectral Imaging.
\end{IEEEbiography}

\vskip -30pt plus -1fil

\begin{IEEEbiography}[{\includegraphics[width=1in,height=1.25in,clip]{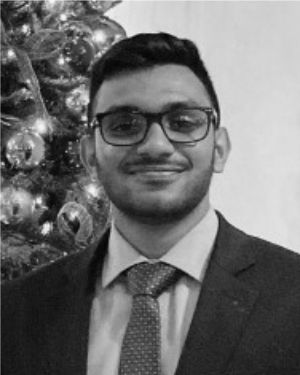}}]{L. Ramanayake} graduated with a second-class honors (upper division) degree in Electrical and Electronic Engineering from University of Peradeniya (2021).  His research interest includes Machine Learning, Spectral Imaging, High Voltage Engineering and Signal Processing. Currently, he is involved in research associated with Hyper-Spectral Imaging for Identification of Probable Mineral Deposits and Application of Multispectral Imaging. 
\end{IEEEbiography}

\vskip -30pt plus -1fil

\begin{IEEEbiography}[{\includegraphics[width=1in,height=1.25in,clip]{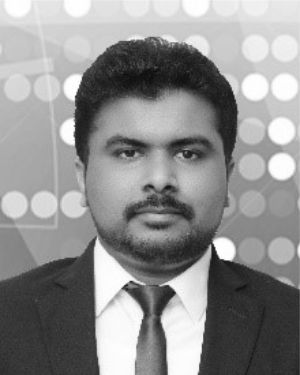}}]{N. Senarath} received a second-class (upper division) honors degree in Electrical and Electronic Engineering from University of Peradeniya (2021). He’s currently working as an instructor at Department of Engineering Mathematics. His research interests include Mathematics, Deep Learning, Digital Communication and Signal Processing. At present, he is involved in research associated with Hyper-Spectral Imaging for Remote sensing and Applications of Multi-Spectral Imaging.
\end{IEEEbiography}

\vskip -30pt plus -1fil

\begin{IEEEbiography}[{\includegraphics[width=1in,height=1.25in,clip]{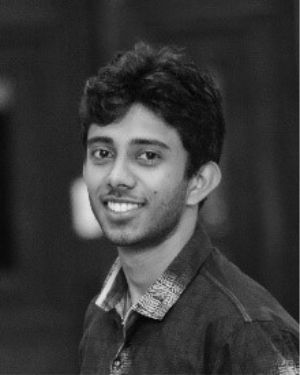}}]{D. Wickramasinghe}, graduated with a second-class honors (upper division) degree in Electrical and Electronic Engineering University of Peradeniya (2021). His research interests include Machine Learning, Multi-Spectral Imaging and Signal Processing. Currently he is involved in researches related to Hyper-Spectral Imaging for Identification of Probable Mineral Deposits, Potable Water Quality Parameter assessing and Identifying Sugar Adulteration in Black Tea through  Multispectral Imaging. 
\end{IEEEbiography}

\end{document}